\documentclass[10pt,letterpaper]{article}
\usepackage[top=0.85in,left=2.75in,footskip=0.75in]{geometry}

\usepackage{amsmath,amssymb}

\usepackage{changepage}

\usepackage{textcomp,marvosym}

\usepackage{cite}

\usepackage{nameref,hyperref}

\usepackage[nopatch=eqnum]{microtype}
\DisableLigatures[f]{encoding = *, family = * }

\usepackage[table]{xcolor}

\usepackage{array}

\newcolumntype{+}{!{\vrule width 2pt}}

\newlength\savedwidth

\usepackage[utf8]{inputenc} 
\usepackage[T1]{fontenc}    
\usepackage{hyperref}       
\usepackage{url}            
\usepackage{booktabs}       
\usepackage{amsfonts}       
\usepackage{nicefrac}       
\usepackage{microtype}      
\usepackage{xcolor}         
\usepackage{amsmath}
\usepackage{amssymb}
\usepackage[inline]{enumitem}
\usepackage{pifont}
\usepackage{makecell} 
\usepackage{caption}
\usepackage{graphicx}
\usepackage{subcaption}
\usepackage{algorithm}
\usepackage{xcolor}
\usepackage{authblk}
\usepackage{multirow}
\usepackage{bbm}
\usepackage{svg}
\usepackage[capitalize]{cleveref}
\crefname{equation}{}{}
\usepackage[section]{placeins}

\usepackage{setspace} 
\raggedright
\usepackage[aboveskip=1pt,labelfont=bf,labelsep=period,justification=raggedright,singlelinecheck=off]{caption}

\makeatletter
\renewcommand{\@biblabel}[1]{\quad#1.}
\makeatother

\usepackage{lastpage,fancyhdr,graphicx}
\usepackage{epstopdf}
\fancyheadoffset[L]{2.25in}
\fancyfootoffset[L]{2.25in}
\newcommand{\indic}[1]{\mathbbm{1}_{[#1]}} 

\newcommand{\methodnamefull}{Molecular-Image Contrastive Learning}
\newcommand{\methodnameabbrev}{MICON}
\newcommand{\contrastivenamefull}{Perturbation-Aware Contrastive Learning of Representations}
\newcommand{\contrastivenameabbrev}{PaCLR}

\begin{document}
\vspace*{0.2in}

\begin{flushleft}
{\Large
\textbf\newline{Integrating chemical structures as treatments improves representations of microscopy images for morphological profiling} 
}
\newline
\\
Yemin Yu\textsuperscript{1\textcurrency},
Emre Hayir\textsuperscript{2},
Neil Tenenholtz\textsuperscript{2},
Lester Mackey\textsuperscript{2},
Ying Wei\textsuperscript{3},
David Alvarez-Melis\textsuperscript{2},
Ava P. Amini\textsuperscript{2},
Alex X. Lu\textsuperscript{2*},
\\
\bigskip
\textbf{1} Department of Computer Science, City University of Hong Kong, Kowloon Tong, Hong Kong
\\
\textbf{2} Microsoft Research, Cambridge, Boston, USA
\\
\textbf{3} Department of Computer Science, Zhejiang University, Hangzhou, China
\\
\bigskip

%

\textcurrency Work primarily done during an internship at Microsoft Research

* Corresponding Author: lualex@microsoft.com

\end{flushleft}
\section*{Abstract}
Recent advances in self-supervised deep learning have improved our ability to quantify cellular morphological changes in high-throughput microscopy screens, a process known as morphological profiling. 
However, most current methods only learn from images, despite many screens being inherently multimodal, as they involve both a chemical or genetic perturbation as well as an image-based readout. 
We hypothesized that incorporating chemical compound structures during self-supervised pre-training could improve learned representations of images from high-throughput microscopy screens. 
We introduce a representation learning framework, \methodnameabbrev{} (\methodnamefull{}), that models chemical compounds as treatments that induce transformations of cell phenotypes. 
\methodnameabbrev{} significantly outperforms classical hand-crafted features such as CellProfiler and existing deep-learning-based representation learning methods in challenging evaluation settings where models must identify reproducible effects of drugs across independent replicates and data-generating centers. 
We demonstrate that incorporating chemical compound information into the learning process provides small, but consistent improvements in performance and that modeling compounds specifically as treatments outperforms approaches that directly align images and compounds in a single representation space. 
Our findings point to a new direction for representation learning in morphological profiling, suggesting that methods should explicitly account for the multimodal nature of microscopy screening data.

\section*{Author summary}
Large-scale microscopy experiments can be useful for many applications, such as discovering new drugs. To analyze these images, scientists will extract representations, series of measurements, that quantify how the shape of cells in these images may change when they're treated with drugs. Recently, deep learning models have been used to learn representations of microscopy images. However, most models only learn from image data alone, despite the fact that many microscopy experiments involve treating cells with drugs, and information about the drug could improve learning. In this work, we propose a new way of training deep learning models that uses both images and the chemical structure of drugs to learn representations for microscopy. Unlike prior proposals, our method models drugs as treatments that alter cells in images. We show that our method learns representations more effective for drug screening applications, and that small, but significant improvements are caused by incorporating information about drugs over just using images alone.

\section*{Introduction}
High-throughput microscopy experiments have become instrumental to drug discovery, guided by the principle that changes in cellular morphology can give insight into drug efficacy and mechanism of action~\cite{lin2020image, ziegler2021morphological}. By coupling scalable fluorescent staining methods like Cell Painting \cite{bray2016cell} with large libraries of small molecules and high-throughput instrumentation, researchers can screen millions of compounds in cell lines. Recent advances in computer vision promise to automate the quantification of increasingly subtle morphological changes in response to these compound treatments \cite{caicedo2017data, pratapa2021image}. The goal of these computational methods is to convert an image into a vector of features, or a representation, that captures cell phenotype, a process called morphological profiling. Representations facilitate the quantitative comparison of phenotypes in microscopy images. For example, to nominate candidate drugs, a researcher might seek to use morphological profiling to identify compounds that yield similar representations --- and therefore similar phenotypes --- to known effective drugs. 

Hand-crafted features, made accessible by the CellProfiler~\cite{carpenter2006cellprofiler} software, have traditionally been used to construct these representations. However, these coarse features often fail to pick up on subtle morphological changes and are confounded by microscopy batch effects that cause irreproducible technical variation. The need for high-quality, robust representations has led researchers to investigate deep-learning based representation learning methods \cite{tang2024morphological}. Since labels for morphological profiling are scarce and frequently unreliable, many representation learning methods train using self-supervised learning, which aims to learn high-quality representations without labeled training data \cite{rani2023self}. For morphological profiling, previous studies have explored various self-supervised frameworks \cite{kim2023self}, including contrastive learning~\cite{chen2020simple, perakis2021contrastive, lin2022incorporating}, masked autoencoding~\cite{he2022masked, kraus2024masked}, and self-distillation~\cite{caron2021emerging, doron2023unbiased}.

While representation learning approaches have shown improvements in sensitivity and robustness relative to hand-crafted features, most methods learn from images exclusively. However, microscopy-based drug screens are inherently multimodal: perturbations, which may be chemical, genetic, or environmental, are used to induce phenotypic changes in cells. Few works have investigated how integrating multimodal information about perturbations, such as the identities of chemical compounds, can improve representation learning. Some weakly supervised methods predict perturbation identity as a target during training \cite{caicedo2018weakly}, but these methods only use the identity of the chemical compound and not any information about its structure. CLOOME~\cite{sanchez2023cloome} and MolPhenix \cite{fradkin2024molecules} integrate chemical structure information through a CLIP (Contrastive Language Image Pre-training)~\cite{radford2021learning} objective to align images and chemical compounds in a unified representation space that enables querying of data from one modality given the other. However, morphological phenotypes and chemical compounds are not directly comparable; for example, depending on cell type, cells may produce different phenotypic responses to the same compound. This may create tensions where a single representation of a chemical compound must be aligned to multiple representations across a dynamic range of phenotypic responses. 

\begin{figure}[!h]
\centering
\includegraphics[width=0.65\textwidth]{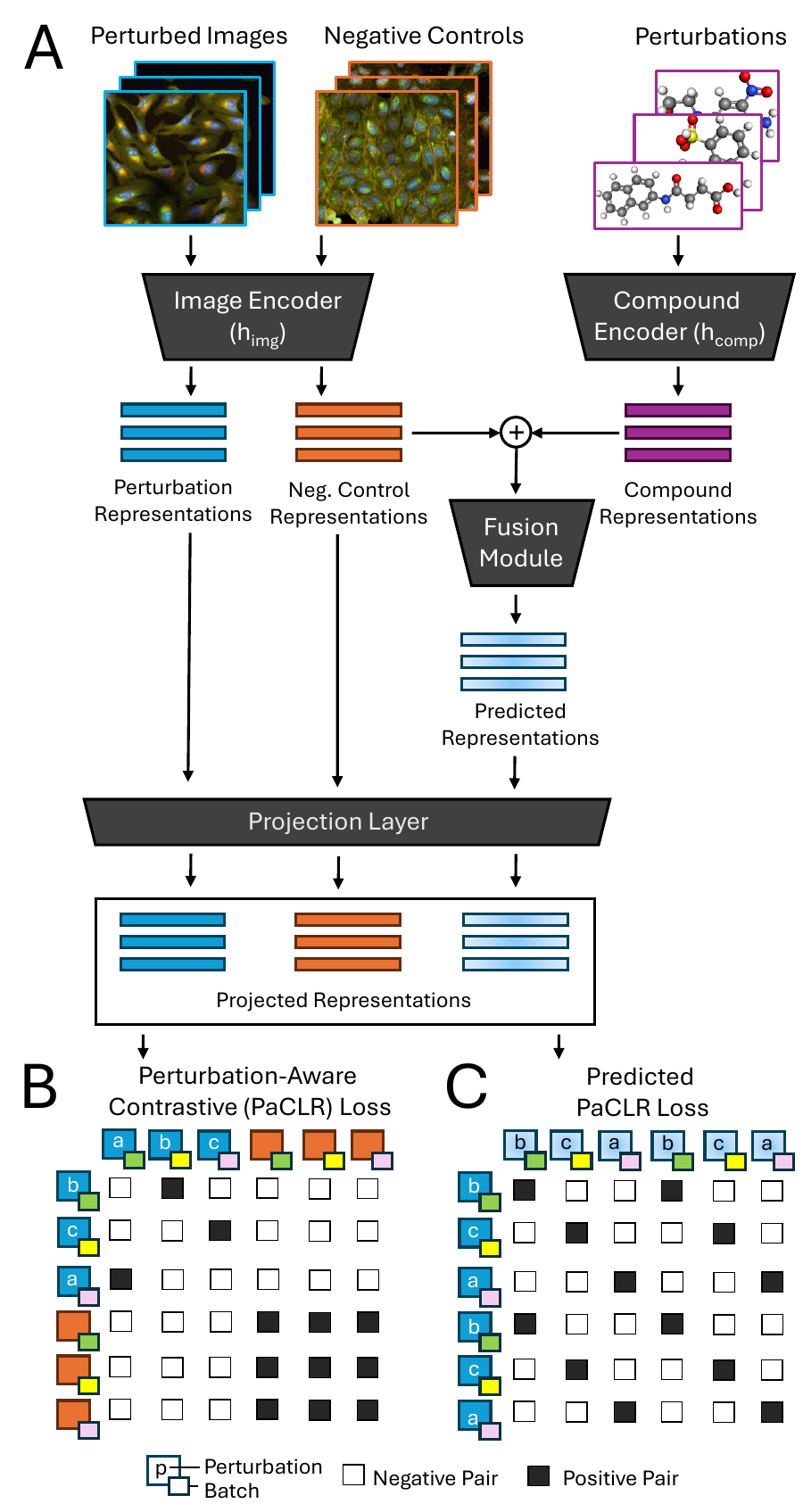}
\caption{\textbf{Summary of the \methodnameabbrev{} framework.} 
\textbf{(A)} An overview of our model. Compound perturbed (blue) images and negative control (orange) images are transformed into representations by an image encoder, while chemical compounds (purple) are transformed into representations by a compound encoder. A fusion module accepts pairs of negative control representations and compound representations and outputs predicted representations (light blue). All representations are passed through a projection layer to form the final projected representations. \textbf{(B)} Summary of the perturbation-aware contrastive (PAC) loss. Perturbed images (blue) and negative controls (orange) from the same batches as the perturbed images (batches denoted by small colored boxes) are sampled, with at least two images for each perturbation. Images treated with the same perturbation (a, b, or c) serve as positive pairs (black) for the \contrastivenameabbrev{} loss. \textbf{(C)} Summary of the \contrastivenameabbrev{} loss on predicted images (abbreviated as Predicted \contrastivenameabbrev{} loss). Perturbation representations (blue) and predicted representations (light blue) using the same compound (a, b, or c) serve as positive pairs (black) for the predicted \contrastivenameabbrev{} loss.}
\label{fig:framework}
\end{figure}

Rather than assuming cell phenotypes in images and chemical compounds to be directly comparable, we instead model chemical compounds as \textit{treatments} that induce transformations of microscopy images to develop \methodnameabbrev{} (\methodnamefull{}), a self-supervised representation learning framework for morphological profiling (Fig~\ref{fig:framework}). Given an image of unperturbed cells and a chemical compound, \methodnameabbrev{} learns how to generate a representation corresponding to a phenotypic transformation of the cells in the image, as if the cells had been treated with the compound.
Using the \methodnameabbrev{} framework, we controllably test the hypothesis that integrating chemical compound information can improve image representation learning in morphological profiling. 
First, we show that \methodnameabbrev{} achieves effective representation learning, substantially improving over both CellProfiler features and previously proposed deep learning methods.
Second, through ablation of \methodnameabbrev{}'s multimodal component, we demonstrate that integrating chemical compound information provides a small but statistically significant improvement in performance relative to learning from images alone.
Finally, supporting our hypothesis that \textit{how} multimodal information is integrated is crucial, we show that modeling chemical compounds as treatments significantly outperforms direct alignment of images and compounds with a CLIP objective. 
Our results advocate for approaching morphological profiling as a multimodal learning problem and provide insight into how to best integrate images and perturbations. 

\section*{Materials and methods}
\subsection*{\methodnameabbrev{}: \methodnamefull{}}

\methodnameabbrev{} is a novel representation learning method for morphological profiling experiments that integrates chemical compounds and images into a multimodal pre-training task (Fig~\ref{fig:framework}). \methodnameabbrev{} models chemical compounds as treatments that induce a transformation in images. Given an image of untreated control cells and a chemical compound, our method transforms the cells in the image as if they had been treated with the chemical compound. In other words, we train \methodnameabbrev{} to simulate the phenotypic effects of chemical compounds, given a chemical compound and an image of untreated cells as input. We perform this transformation directly in the image representation space, using contrastive losses to align predicted and real representations (Fig~\ref{fig:framework}A). 
We next describe the two components of our representation learning approach: a perturbation-aware contrastive learning objective that learns faithful representations of phenotypes in real microscopy images and a predicted perturbation-aware contrastive objective that aligns predicted representations and perturbation representations.

\subsubsection*{\contrastivenamefull{}}

As our method relies on alignment of real and predicted perturbation representations, we must first ensure that representations of real phenotypes are robust and capture the phenotypic effects of perturbations. However, learning such a representation is challenging because of microscopy batch effects, which induce differences in images that stem from irreproducible technical variation \cite{arevalo2024evaluating}.

Batch effects are known to be a major confounder even in replicates at the same data-generating institution \cite{caicedo2017data}. Our work leverages a dataset (cpg0016, see Datasets and Evaluation) where images are supplied by multiple sources, which introduces further differences between experiments (e.g., images may be collected under different microscopes) and thus even more severe microscopy batch effects \cite{arevalo2024evaluating}. If representations of real phenotypes are not robust, learning to predict the phenotypic impact of perturbations may result in spurious learning of microscopy batch effects, not the perturbation. To learn representations robust to microscopy batch effects, we design a \contrastivenamefull{} (\contrastivenameabbrev{}) objective, which leverages metadata about perturbation and microscopy batches to sample positive and negative pairs for contrastive learning.

\newcommand{\img}{\mathbf{x}} 
\newcommand{\rep}{\mathbf{t}} 
\newcommand{\perturb}{\mathbf{p}} 
\newcommand{\batch}{\mathbf{b}} 
\newcommand{\imgenc}{\mathbf{h}_{\mathrm{img}}} 
\newcommand{\compenc}{\mathbf{h}_{\mathrm{comp}}} 

We first consider a more standard self-supervised contrastive learning objective, i.e. SimCLR \cite{chen2020simple}. SimCLR trains using self-augmented pairs of images. In this setting, an image $\img$ is encoded by an image encoder $\imgenc$, then passed through a projection head $g$ to obtain a projected representation $\mathbf{t} = g(\imgenc(\img))$. 
Given $N$ images with projected representations $\mathbf{t}_1, \cdots, \mathbf{t}_N$, a contrastive loss can be used to pull together the representation of an image and its augmentations (positive pairs), while repelling representations of dissimilar images (negative pairs). For a given image index $i$, we let $j$ index a differently augmented version of the same image and $k$ iterate over all images other than $i$ in the set (where $k \neq j$ are assumed to be dissimilar to $i$). The contrastive loss is then given by

\begin{equation*}
    \mathcal{L}_{i, j} = - \log \frac{\texttt{exp}(\tau^{-1}\texttt{sim}(\mathbf{t}_i, \mathbf{t}_j))}{\sum _{k=1}^{N} \indic{i \neq k} \text{exp}(\tau^{-1}\texttt{sim}(\mathbf{t}_i, \mathbf{t}_k))},
\end{equation*}
where $\tau$ is the temperature scale hyperparameter for the softmax function and \texttt{sim} is a similarity measure (in our work, the cosine distance, consistent with standards established by SimCLR \cite{chen2020simple}).

This naive contrastive strategy poses two problems in the context of morphological profiling. The first is class collision. In the above setting, all other images in a training batch are considered negative samples. In high-throughput imaging however, some images in the same training batch may share the same perturbation and thus carry related morphological changes that we wish to capture. Second, augmented pairs of images are still likely to share microscopy batch effects, which makes a standard contrastive strategy prone to learning confounding technical variation, instead of focusing on the effect of the perturbation alone.

By leveraging metadata about the perturbation and microscopy batch to sample pairs for contrastive learning, the \contrastivenameabbrev{} loss addresses both of these issues (Fig \ref{fig:framework}B). First, we use metadata about perturbations to define positive pairs. Instead of using different augmented views of a single instance, we use images of the same perturbation as positive pairs. The underlying assumption is that, if the network learns to align two images of the same perturbation, the shared signal that is learned should be the biological signal, rather than confounding microscopy batch effects (since these images are drawn uniformly from the dataset, positive pairs can come from different microscopy batches, although this is not strictly enforced). Second, we use metadata about microscopy batch to sample negative control images (which are not treated with any compound, but rather a control treatment of only the solvent in which compounds are dissolved in). This means there are samples that share microscopy batch effects, but not perturbation, serving as negative examples, further encouraging the model to be robust to microscopy batch effects. We sample negative control images such that they come from the same microscopy plate as perturbed images or from the same microscopy batch as a fall-back for the rare case that there are no viable control images on the same plate (see Datasets and Evaluation for a description of the structure of microscopy experiments). In addition to image instances perturbed by different compounds, these negative control images act as negative pairs in training, which further discourages the \methodnameabbrev{} model from learning about microscopy batch effects, as negative pairs are enforced to come from the same microscopy batches as perturbed images.

\newcommand{\imgset}{\mathcal{I}}
\newcommand{\treatset}{\mathcal{T}}
\newcommand{\controlset}{\mathcal{C}}
The above is implemented through sampling of batches during training. Given an image $\img$, let $\perturb$ be its perturbation, and $\batch$ be its microscopy batch identifier. We first sample $T$ perturbed images from the entire dataset to form the subset $\treatset_1 = \{\img_{1}^{\treatset_1},...,\img_{T}^{\treatset_1}\}$. Next, for each image in $\treatset_1$, we sample a second distinct image from the dataset with the same perturbation to form the subset $\treatset_2 = \{\img_{1}^{\treatset_2},...,\img_{T}^{\treatset_2}\}$, so $\perturb_{j}^{\treatset_1}= \perturb_{j}^{\treatset_2}$ but $\img_{j}^{\treatset_1}\neq \img_{j}^{\treatset_2}$. Finally, we sample $C$ negative control images from the dataset to form the subset $\controlset=\{\img_{1}^{\controlset},...,\img_{C}^{\controlset}\}$, such that their microscopy batch identifiers match as many of the perturbed images ($\treatset_1 \cup \treatset_2$) as possible given the size of the subset: $\batch_{j}^{\controlset} \in \{\batch_{1}^{\treatset_1},...,\batch_{T}^{\treatset_1}\} \cup \{\batch_{1}^{\treatset_2},...,\batch_{T}^{\treatset_2}\}$, and $|\{\batch_{1}^{\controlset},...,\batch_{C}^{\controlset}\}| = \min(|\{\batch_{1}^{\treatset_1},...,\batch_{T}^{\treatset_1}\} \cup \{\batch_{1}^{\treatset_2},...,\batch_{T}^{\treatset_2}\}|, C)$. Putting these together, the training batch is the union of these three subsets $\treatset_1 \cup\treatset_2 \cup \controlset$.

Finally, we extract a representation $\rep_i = g(\imgenc(\img_i))$ from each image $\img_i$ using our image encoder $\imgenc$. 
With this notation, we define the \contrastivenameabbrev{} loss, 
\begin{align}\label{eq:orig-bac-loss}\mathcal{L}^{\text{real}}_{\text{PaCLR}}  = -\frac{1}{N}\sum_{i=1}^N\frac{1}{\sum_{k=1}^N\indic{\perturb_i = \perturb_k, i\neq k}} \sum _{j=1}^{N} \indic{\perturb_i = \perturb_j, i\neq j}\log \frac{\texttt{exp}(\tau^{-1}\texttt{sim}(\mathbf{t}_i, \mathbf{t}_j))}{\sum _{k=1}^{N} \indic{i \neq k} \texttt{exp}(\tau^{-1}\texttt{sim}(\mathbf{t}_i, \mathbf{t}_k))}.
\end{align}

\contrastivenameabbrev{}'s integration of metadata about perturbation into representation learning is an idea also explored by several previous works, but our implementation of this idea has several key differences. Compared to Set-DINO \cite{yao2024weakly} and WS-DINO \cite{cross2022self}, we use images treated with the same perturbation as positive pairs for contrastive learning, instead of enforcing similarity in a student-teacher set-up. Compared to weakly supervised proposals \cite{caicedo2018weakly, moshkov2024learning}, we use metadata about perturbations for sampling pairs, rather than as predictive targets. Finally, we note that unlike many other prior works which only use perturbed images \cite{cross2022self, sanchez2023cloome}, we explicitly include negative control images. We incorporate negative controls because, due to their lack of perturbation treatment, they primarily represent phenotypic changes caused by microscopy batch effects and thus enable disentanglement of microscopy batch effects from perturbation effects.

\subsubsection*{Predicted Perturbation-Aware Contrastive Objective}

To incorporate chemical compound information into \methodnameabbrev{}'s learning objective, we first encode chemical compounds into representations using a compound encoder $\compenc$ (Fig~\ref{fig:framework}A). A fusion module $F$ combines the resulting representations of chemical compounds with the representation of a negative control image. The fusion module predicts representations reflecting if cells were to be treated with a given perturbation (Fig~\ref{fig:framework}A). To ensure these predicted representations are accurate, we align predicted representations with real representations of images of cells actually treated with the compounds using a variant of the perturbation-aware contrastive loss. Unlike the original \contrastivenameabbrev{} loss (Equation \cref{eq:orig-bac-loss}), the \contrastivenameabbrev{} loss on the predicted images (abbreviated as Predicted \contrastivenameabbrev{} loss in Fig \ref{fig:framework}) is calculated using only perturbation representations, i.e., excluding the negative controls (Fig \ref{fig:framework}C). However, the negative control images used to generate predicted representations in the fusion module are sampled such that they come from the same microscopy batches as the real perturbation representations.

Formally, we use a fusion model $F$ to combine the representation 
of a chemical compound $\mathbf{\tilde{p}}$ and the representation 
of a negative control image $\img^{c}$ into a predicted representation $\mathbf{\tilde{t}} = F(g(\imgenc(\img^{c})), \compenc(\mathbf{\tilde{p}}))$.
Then, given $N/2$ perturbation representations and their perturbation treatments, denoted as the tuple $(\mathbf{t}_i,\perturb_i)$, and $N/2$ predicted representations and their perturbation treatments, denoted as the tuple $(\mathbf{\tilde{t}}_j,\mathbf{\tilde{p}}_j)$, we define the following \contrastivenameabbrev{} loss on the predicted images over the predicted representations and the real perturbation representations: 
\begin{equation*}
\mathcal{L}^{\text{pred}}_{\text{PaCLR}} = -\frac{2}{N}\sum_{i=1}^{N/2}\frac{1}{\sum_{k=1}^{N/2}\indic{\perturb_i = \mathbf{\tilde{p}}_k}} \sum _{j=1}^{N/2} \indic{\perturb_i = \mathbf{\tilde{p}}_j}\log \frac{\text{exp}(\tau^{-1}\text{sim}(\mathbf{t}_i, \mathbf{\tilde{t}}_j))}{\sum _{k=1}^{N/2} \text{exp}(\tau^{-1}\text{sim}(\mathbf{t}_i, \mathbf{\tilde{t}}_k))}.
\end{equation*}

The final loss for \methodnameabbrev{} combines the \contrastivenameabbrev{} objectives on the real versus predicted images. We weight these loss components equally:
\begin{equation*}
\mathcal{L} = \mathcal{L}^{\text{real}}_{\text{PaCLR}} + \mathcal{L}^{\text{pred}}_{\text{PaCLR}}
\end{equation*}

\subsection*{Datasets and Evaluation}

\subsubsection*{Training Dataset}
\label{sec:data}

We leveraged a curated subset of the JUMP Cell Painting Consortium's (JUMP-CP) cpg0016 dataset, a large-scale dataset cell painting microscopy images of drug-treated U2-OS cells \cite{chandrasekaran2023jump}. We chose cpg0016 as it consists of data from 12 data-generating sources, enabling us to test robustness to microscopy batch effects more extensively. 

Each source collects images of cells cultured on \textit{plates}, which are sub-divided into 384 or 1536 \textit{wells}, where each well is capable of containing an independent experiment treated with a different perturbation. A robot-controlled microscope takes images of the experiment in each well, typically multiple images of different parts of the well, where each distinct image of the same well is a \textit{field of view (FOV)}. Plates are further organized into \textit{batches}, which we term a \textit{microscopy batch} to distinguish from a batch of data for model training, where plates in the same microscopy batch are imaged on the same day. While performing the same perturbation experiment across different wells, plates, microscopy batches, or sources can all generate microscopy batch effects, microscopy batch effects are generally considered more pronounced across different microscopy batches or sources, as opposed to different wells on the same plate, or different plates in the same microscopy batch \cite{arevalo2024evaluating} (supporting this, we show the average distance between negative control CellProfiler features across batches and sources in JUMP-CP in \nameref{S1_Fig}). Further details on the dataset and data generation setup can be found in the original JUMP-CP paper \cite{chandrasekaran2023jump}.

We curated several kinds of plates and wells from the cpg0016 dataset for training and evaluating our models. First, we curated positive control wells (abbreviated as POS-CTL in this work) from the entire dataset; these reflect 8 compounds selected to have the most distinct morphological changes in a pilot study. Most plates in cpg0016 have 4 well replicates of each of these 8 positive controls. Second, we curated JUMP-Target-2-Compound plates (abbreviated as Target-2 in this work), which image 301 compounds selected as likely to induce phenotypic effects in cells, and are measured in replicate by every source. Third, we curated a subset of 184 compounds from the Compound plates of cpg0016 (see Data Stratification and Compound-Replicate Matching for further details). Finally, every plate includes negative control wells treated with Dimethyl Sulfoxide (DMSO), the solvent used for all compound perturbations, which acts as a control where no compound is dissolved in DMSO. We curate negative controls from all plates from which we sampled POS-CTL, Target-2, and Compound wells. Together, our dataset comprises 87,282 of the 1,096,069 wells in cpg0016. Note that for our loss, we consider a well to be an image $\img_i$, and sample a random FOV for the well. This has the impact that no FOVs from the same well are ever within the same batch of data during training of our method.

\subsubsection*{Data Stratification and Compound-Replicate Matching}
\def\nntrain{retrieval}
\def\nntest{query}
\label{sec:stratification}
We purposed cpg0016 for both training, and for compound-replicate matching evaluations, a standard strategy for evaluating if representations robustly identify phenotypes in the presence of microscopy batch effects \cite{caicedo2017data, chen2024chammi}. The principle underlying compound-replicate matching is that cells treated with the same perturbations should have similar phenotypes, and therefore have similar representations, even if experiments originate from different microscopy batches or sources. Compound-replicate matching is typically implemented through nearest-neighbor retrieval \cite{chen2024chammi, ljosa2013comparison, ando2017improving, janssens2021fully}, which means for data stratification, in addition to defining \textit{training} and \textit{validation} datasets for representation learning, we also need to define compound-replicate matching \textit{\nntrain{}} and \textit{\nntest{}} datasets for evaluation. During evaluation, we retrieve the nearest neighbor in the \nntrain{} dataset for each well from the \nntest{} dataset, and measure whether these share the same perturbation treatment. There should be no overlap between the \nntest{} dataset and each of the model training and validation datasets, to avoid overfitting to images seen during representation learning.

\begin{figure}[!h]
\centering
\includegraphics[width=\textwidth]{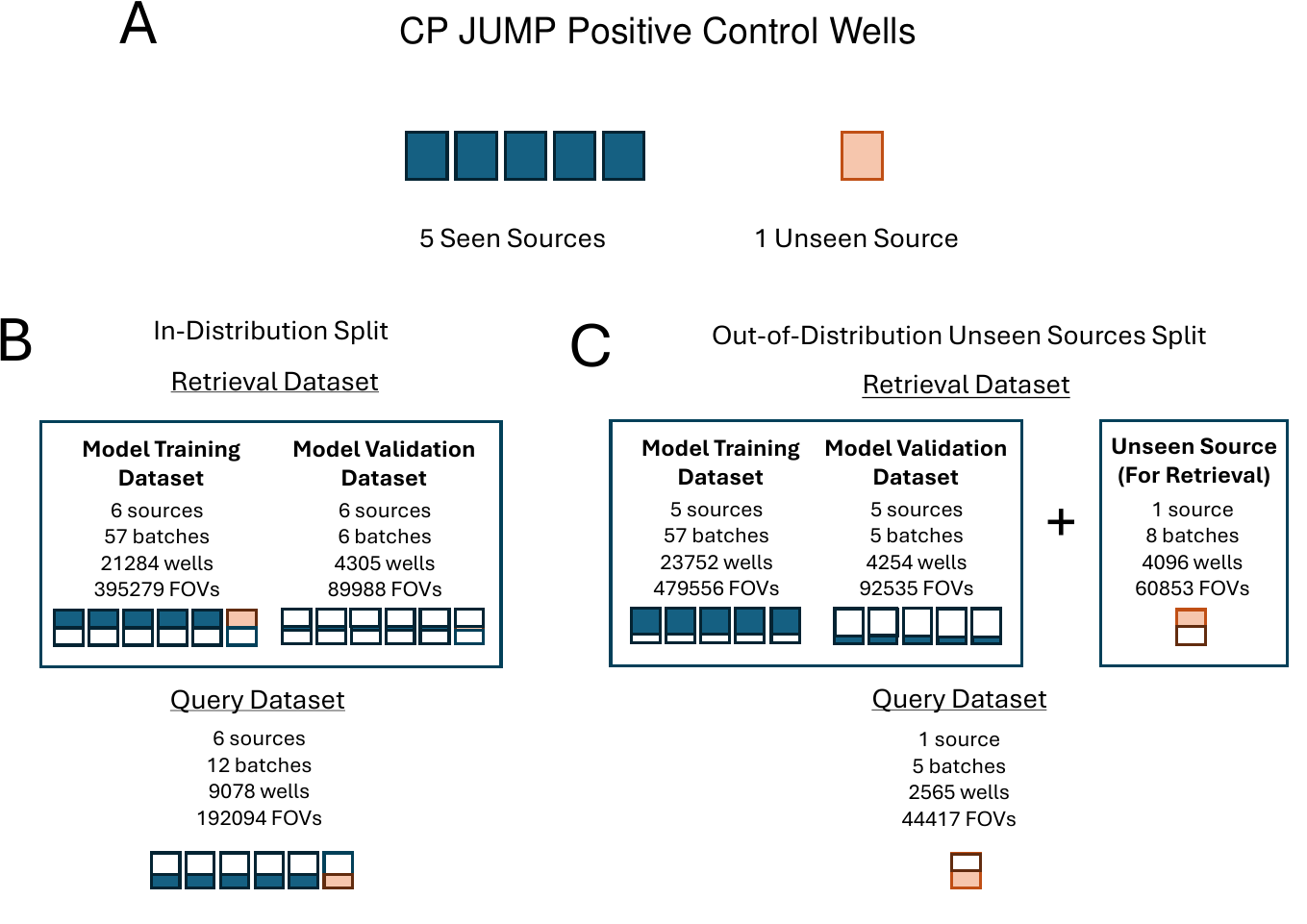}
\caption{\textbf{Stratification of positive control wells into in-distribution and out-of-distribution datasets.} \textbf{(A)} POS-CTL wells from a total of six sources are used. For our out-of-distribution evaluation, five sources are designated as seen (blue boxes) during model training, and 1 source is designated as unseen (orange boxes). In panels B and C, filled parts of the boxes indicate data from each source in that split, while unfilled (white) parts of the boxes indicate held-out data (note that filled/unfilled ratios are not proportionately scaled to split ratios, and are just intended to illustrate where splits are disjoint). \textbf{(B)} The in-distribution evaluation trains the representation learning model on a subset of microscopy batches from all six sources, holding out one batch from each source for validation. The training and validation dataset are used as the \nntrain{} dataset for compound-replicate matching, with unseen microscopy batches from all six sources used as the \nntest{} dataset. \textbf{(C)} The out-of-distribution evaluation trains and validates the representation learning model on the five seen sources. For evaluation, the \nntrain{} dataset is the training and validation data plus a subset of microscopy batches from the unseen source, and the \nntest{} dataset is a disjoint subset of microscopy batches from the unseen source.
}
\label{fig:pos_control_stratification}
\end{figure}

\begin{figure}[!h]
\centering
\includegraphics[width=\textwidth]{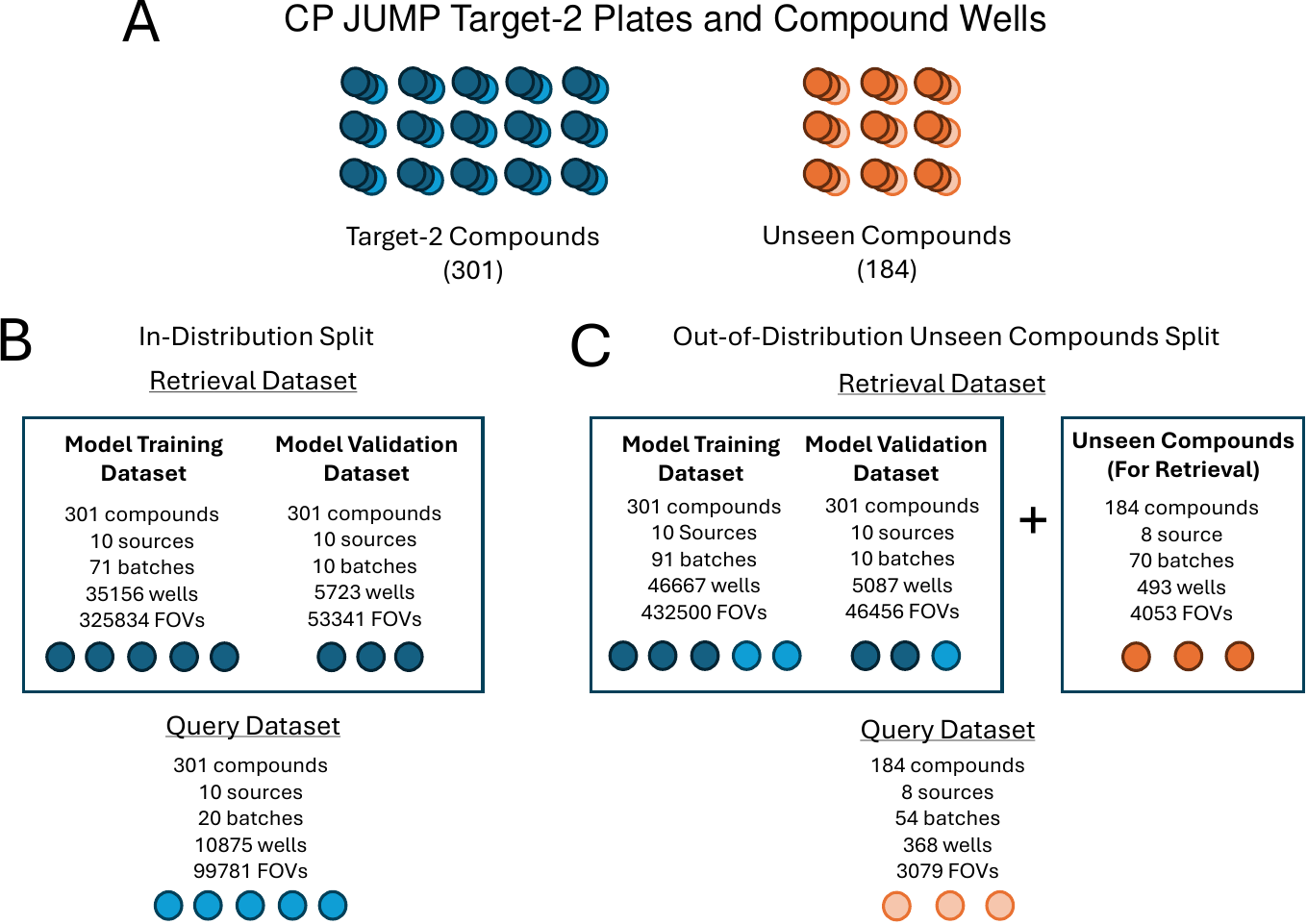}
\caption{\textbf{Stratification of Target-2 and Compound plates into in-distribution and out-of-distribution datasets.} \textbf{(A)} The 301 Target-2 compounds (blue) are designated as seen compounds for training representation learning models, while 184 unseen compounds (orange) are sampled from the Compound plates and held out. \textbf{(B)} The in-distribution evaluation trains the representation learning model on 80\% of microscopy batches from all Target-2 plates, reserving 20\% of microscopy batches for the \nntest{} dataset. \textbf{(C)} The out-of-distribution evaluation trains the representation learning model on all microscopy batches from the Target-2 plates. The training dataset is combined with 493 wells for 184 unseen compounds to form the \nntrain{} dataset and evaluated on a \nntest{} dataset of a disjoint set of 368 wells of held-out, unseen compounds from the Compound plates.}
\label{fig:tgt2_stratification}
\end{figure}

Our data stratification strategy is motivated by two guiding use cases for morphological profiling: first, the identification of reproducible biological changes in unseen data for sources and compounds seen during training, which we refer to as ``in-distribution~(ID) generalization''; and second, the identification of these changes in unseen data for sources and compounds \textbf{not} seen during training, which we refer to as ``out-of-distribution~(OOD) generalization''. The first use case captures scenarios such as those when a source trains a model on its own data and then seeks to evaluate it on newly-produced data. The second use case captures settings such as those when a distinct source seeks to use a pre-existing model, trained on data from other sources, to analyze their data, or when a source seeks to use a pre-existing model on new compounds not seen in training. To capture ID and OOD performance for \textit{both} unseen sources and unseen compounds, we produce four distinct data splits on different subsets of data within the cpg0016 dataset (Fig~\ref{fig:pos_control_stratification},~\ref{fig:tgt2_stratification}).

We used the POS-CTL dataset to evaluate model generalization to unseen sources (Fig~\ref{fig:pos_control_stratification}A). We stratified the dataset by source in two ways to test ID and OOD performance (Fig \ref{fig:pos_control_stratification}B-C, with metadata about splits available in \nameref{S9_Table}). For the ID setting (Fig~\ref{fig:pos_control_stratification}B), we used the 6 cpg0016 sources that have at least 10 POS-CTL plates. We trained representation learning models on a subset of data from all 6 sources stratified by microscopy batch, and we held out one batch from each source as validation data for model selection. This training dataset is recycled as the \nntrain{} dataset, but we reserved unseen batches from all 6 sources as the \nntest{} dataset (Fig~\ref{fig:pos_control_stratification}B; \nameref{S9_Table}). We split microscopy batches into the \nntrain{} dataset versus the \nntest{} dataset at a 85-15 ratio. 
For the OOD setting (Fig~\ref{fig:pos_control_stratification}C), we used the same 6 cpg0016 sources as before, but created a training dataset containing data from just 5 of the 6 sources. The \nntrain{} dataset includes data from the training dataset, but also a subset of microscopy batches from the unseen source (50\% of batches). We added the unseen source to the \nntrain{} dataset because this allows us to quantify two settings for a source not included in training data: first, if the source uses the pre-trained model to compare between images generated internally, and second, if the source uses the pre-trained model to compare their images to another different source (NSB and NSS respectively, see Compound-Replicate Matching Implementation). The \nntest{} dataset is a disjoint 50\% of microscopy batches from the unseen source. Dataset stratification is randomized over 3 random seeds, and this is a source of uncertainty in our error bars (in addition to randomizing weights, where applicable for the method).

We used the Target-2 dataset to evaluate generalization to unseen compounds. We stratified the Target-2 dataset to test ID performance, and used a mixture of Target-2 and Compound plates to test OOD performance (Fig \ref{fig:tgt2_stratification}, with metadata about splits available in \nameref{S10_Table}). For the ID setting (Fig~\ref{fig:tgt2_stratification}B), we reserved 20\% of the microscopy batches as the \nntest{} dataset. The remaining 80\% of batches are split into training and validation datasets (holding out one batch from each source for validation), and additionally used as the \nntrain{} dataset. For the OOD setting (Fig~\ref{fig:tgt2_stratification}C), we trained representation learning models using all batches from the Target-2 plates. To evaluate, we extended the dataset by identifying compounds from the Compound plates that elicited strong phenotypic effects across sources. For each source we ranked compounds in the Compound plates based on the replicate-averaged distance between the CellProfiler~\cite{mcquin2018cellprofiler} image features for the compound versus features for the negative control spatially closest to the well in the plate map. We then nominated compounds in the top 10\% of greatest distances for at least four sources, yielding a total of 184 compounds. All of these compounds have disjoint mechanisms of action (MoAs) with the Target-2 compounds. Our \nntrain{} dataset combines the training dataset with a subset of wells (493 wells) for the unseen compounds, while the \nntest{} dataset contains 368 unseen wells (2 wells per compound). This stratification scheme means that the 184 unseen compounds can be matched to 485 compounds (both the 301 compounds seen and the 184 compounds unseen), making the compound-replicate matching task more difficult and penalizing models if they overfit to compounds seen in training.

\subsubsection*{Compound-Replicate Matching Implementation}
\label{sec:evaluation}

To implement compound-replicate matching, we identified the nearest neighbor in the \nntrain{} dataset, using the cosine distance of the representation, for each well in the \nntest{} dataset. We reported the accuracy of this retrieval, defined as the fraction of wells where the nearest well in the \nntrain{} dataset is treated with the same perturbation. As each well is typically imaged with multiple FOVs, we averaged the representations of all FOVs to form a well-level representation. To test if representations are robust to microscopy batch effects caused by differences in microscopy batch and source, we further restricted matching to the nearest neighbor in a different microscopy batch (Not in Same Batch - NSB) or source (Not in Same Source - NSS).

\subsubsection*{Mechanism of Action Enrichment}

In addition to compound replicate matching, we also evaluated models on mechanism of action (MoA) enrichment. We use the BBBC036 dataset, previously proposed for this evaluation by \cite{moshkov2024learning}. We further implement this evaluation using images from the JUMP-CP compound dataset. MoA labels were obtained from the Drug Repurposing Hub \cite{drug_repurposing_hub}, and for this task only compounds that are labeled in the Drug Repurposing Hub with a corresponding MoA was used. A query compound was excluded from the task if there was no other compound with a shared MoA. This evaluation aims to assess if compounds with biologically similar effects are clustered in representation space, and measures retrieval of distinct compounds with similar MoAs. In this evaluation, all other compounds are ranked using a query compound. The top 1\% of matching compounds is evaluated for enrichment in compounds with the same MoA as the query compound, relative to the other 99\% of compounds in the ranked list, reported using the odds-ratio (or fold-of-enrichment). Finally, the odds-ratio is averaged across all query compounds (see \nameref{S1_Text} for a full description of how metrics are calculated). As the BBBC036 is distinct from JUMP CP, MoA enrichment evaluation measures performance on a fully held-out dataset (BBBC036), as well as the more in-distribution JUMP-CP dataset.

In addition to reporting the average fold-of-enrichment following \cite{moshkov2024learning}, we reported the proportion of compounds with a significant enrichment (below a p-value of 0.05 for the permutation test used to calculate enrichment), as a more robust metric. Many MoAs have few compounds associated with them, so for individual MoAs, all compounds with the same MoA may be ranked in the top 1\% of retrieved hits. This leads to an infinite odds ratio, which the authors deal with by imputing the size of the background set when this happens. However, this imputation strategy can significantly inflate the average fold-of-enrichment metric as the size of the background set (1534 for BBBC036 and 941 for JUMP-CP) is typically much higher than the typical finite fold-of-enrichment (on average 7 to 11 fold). This can lead to exaggerated differences between methods with minimal differences in behavior (e.g. if one method ranks all matching compounds within the top 1\%, and another compound ranks all but one matching compound in the top 1\%).

\subsection*{Implementation}
\subsubsection*{Image Preprocessing}

To remove systematic variation in pixel intensity from uneven illumination in the field of view (FOV), we applied illumination correction~\cite{singh2014pipeline}. Each image was corrected by dividing all pixel intensities by the illumination correction function, calculated by taking a median filter on the average image across all FOVs in a plate. Consistent with previous representation learning studies on cell painting images \cite{moshkov2024learning}, to reduce the disk size of the dataset, images are compressed from 16-bit TIFF files to 8-bit PNG files, rescaling each channel independently between the 0.05th and 99.95th percentile of pixels intensity-wise. To further save disk space and improve I/O speed, we take a crop of 50\% of the image height and width from the center of each image. For data augmentation and to allow large batch sizes, the processed images are further re-sized to 224x224 resolution during training. Following previous works \cite{ando2017improving, arevalo2024evaluating, way2022morphology}, we also evaluated the effect of spherizing representations: we first rescaled features with mean absolute deviation (MAD) normalization, and then we applied spherizing, which transforms the feature space such that negative control representations have an identity covariate matrix.

\subsubsection*{Architecture and Training Details}
\label{sec:training}

We use ResNet101~\cite{he2016deep}, pre-trained on ImageNet, as our image encoder $\imgenc$, following previous works showing that ImageNet features are effective at morphological profiling~\cite{kraus2016classifying,pawlowski2016automating,godinez2017multi,cross2022self}. We replaced the final fully-connected layer with a new randomly initialized layer of hidden dimension size 256 and fine-tune only this new layer, with the pre-trained encoder frozen during training. Since models pre-trained on ImageNet typically expect 3-channel inputs, we preprocessed 5-channel microscopy images by converting each channel separately into a 3-channel image by duplicating the image 3 times along the channel axis. Each microscopy channel is separately preprocessed in this manner and inputted into the encoder to output an encoding of dimension $256$. We concatenate the outputted encodings from the 5 original microscopy channels to obtain a final image representation with dimension $256\times 5=1280$.

To encode chemical compounds, we represented them as ECFP4 fingerprints with a radius of 2~\cite{rogers2010extended}, which represents molecular structures as circular neighborhoods of atoms. The ECFP4 fingerprint transforms a molecular structure to a bit-vector embedding which captures patterns within a two-bond radius around each atom, enabling efficient comparison of molecular structures. The compound encoder, $\compenc$, is trained from scratch. $\compenc$ contains 4 fully-connected hidden layers of dimensionality 2048 each and interleaved with batch normalization and ReLU activations, receives as input the ECFP4 fingerprint, and outputs a molecular encoding of dimension 2048. We also tested starting from a pre-trained 3D molecular encoder, UniMol~\cite{zhou2023uni}, but observed in initial experiments that this did not improve performance. 

The fusion module accepts as input a concatenation of a negative control image embedding and a molecule embedding. The model is implemented as a two-layer multi-layer perceptron with leaky-ReLU activations. Finally, all image representations (i.e. both representations of real images and predicted representations) are passed through a projection head $g$, also implemented as a two-layer multi-layer perceptron with leaky-ReLu activations. Together, these components (the image encoder $\imgenc$, the chemical compound encoder $\compenc$, the fusion module $F$, and the projection head $g$) are combined into the overall \methodnameabbrev{} framework (Fig \ref{fig:framework}). The entire model is optimized end-to-end with the overall \methodnameabbrev{} loss, with the exception of the frozen components of the image encoder. 

We trained using the AdamW optimizer~\cite{kingma2014adam} using a ReduceLROnPlateau learning rate scheduler using batch size $64$ for $30$ epochs on two NVIDIA A100-40G graphic cards. We oversampled negative controls to form half of the training batch for our POS-CTL models and one eighth of the training batch for our Target-2 models. We perform a linear warm-up for $2000$ steps. All models are trained 3 times from random seeds. We saved checkpoints at every 2000 steps, and use the best-performing checkpoint on the validation dataset for testing. Detailed hyper-parameters and architectural details can be found in \nameref{S11_Table}. 

\subsubsection*{Baselines and Ablations}

We compared \methodnameabbrev{} to the following baselines and ablations: hand-crafted, image-based features (``CellProfiler''); an image-only, self-supervised contrastive learning model (``SimCLR''); and a CLIP-based multi-modal model that aligns image and chemical compound representations (``CLIP''). 

CellProfiler~\cite{mcquin2018cellprofiler} features -- which are hand-designed, image-based texture, shape, and intensity features -- were provided by and downloaded from the JUMP-CP dataset \cite{chandrasekaran2023jump}. Models trained using SimCLR \cite{chen2020simple} provided an image-only, self-supervised contrastive learning baseline to validate our strategy of integrating chemical compound information in a multi-modal framework. To specifically evaluate if \methodnameabbrev{} treating perturbations as treatments was effective, we trained multi-modal baseline models using CLIP \cite{radford2021learning}, which directly aligns image and chemical compound representations in a shared embedding space. A CLIP-based approach for multimodal learning between microscopy images and chemical compounds was previously proposed by CLOOME ~\cite{sanchez2023cloome} and extended in MolPhenix~\cite{fradkin2024molecules}. As these works use different architectures, datasets, and pre-trained models, they do not provide appropriate baseline controls to \methodnameabbrev{}. We trained a CLIP baseline model using the same dataset and encoder architectures as \methodnameabbrev{} to enable direct comparison of the method itself, instead of transferring these models, which could result in disparities in performance from these other factors. Thus, we refer to the resulting baseline model as ``CLIP.''

Both the SimCLR and CLIP models use the same architectures, dataset, and stopping criterion as \methodnameabbrev{} listed in the Archiecture and Training Details, when applicable to the model (i.e., SimCLR and CLIP also use ResNet101 pre-trained on ImageNet, while CLIP also uses our ECFP4-based compound encoder trained from scratch). All models are trained $3$ times with distinct random seeds. We also test features extracted from the ResNet101 pre-trained encoder that we initialized SimCLR and CLIP with, with no further training, to confirm that SimCLR and CLIP improve over this initialization.

In addition to SimCLR and CLIP, we also fine-tuned DINOv3 models on our datasets, a state-of-the-art self-supervised vision method. We initialize the VIT-Base model trained by \cite{simeoni2025dinov3}, and fine-tuned on our dataset following their method, for $30$ epochs with ReduceLROnPlateau learning rate scheduler starting with a learning rate of $10^{-3}$. The rest of the optimizer settings are identical to the setup in Architecture and Training Details. We also test a DINOv3 model without fine-tuning to confirm that fine-tuning induces improvements in performance.

Finally, we extracted features from OpenPhenom-S/16 \cite{kraus2024masked}, a foundation model for cell painting trained using masked autoencoding, 

\subsubsection*{Code and Data Availability}
Code to reproduce our analyses and model is available at \url{github.com/microsoft/MICON}. 
Model weights and representations from our models are available at \url{https://huggingface.co/datasets/Yemin/MICON/tree/main}. As the image datasets are readily available from the JUMP Cell Painting Consortium \cite{chandrasekaran2023jump}, and too large to re-upload, we instead provide scripts to reproduce our splits and preprocessing in our GitHub repository at \url{github.com/microsoft/MICON}. 

\section*{Results}
\begin{figure}[!h]
\centering
\includegraphics[width=\textwidth]{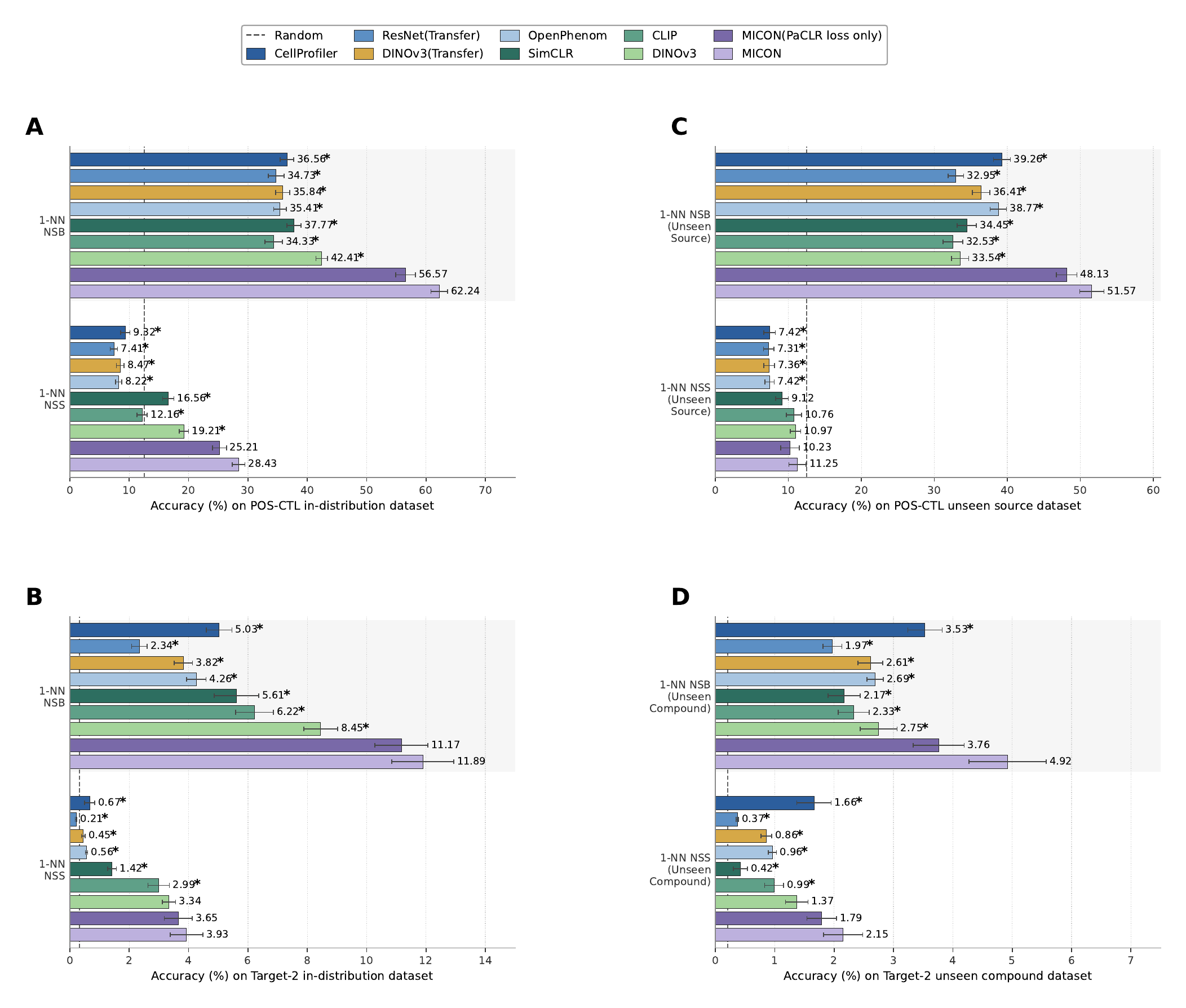}
\caption{\textbf{Compound-replicate matching accuracy for representation learning methods across evaluation settings.} Retrieval accuracies on the \textbf{(A)} POS-CTL ID \nntest{} dataset, \textbf{(B)} Target-2 ID \nntest{} dataset, \textbf{(C)} POS-CTL OOD \nntest{} dataset (unseen source), and \textbf{(D)} Target-2 OOD \nntest{} dataset (unseen compounds). NSB designates retrieval of the nearest neighbor not in the same batch; NSS designates retrieval of the nearest neighbor not in the same source. $n$=3 models trained and evaluated with different dataset stratifications and different random seeds; the mean accuracy is labeled, and the error bars represent standard deviation. Asterisk indicates that \methodnameabbrev{} significantly outperforms the baseline, defined as $p<0.05$ on an unpaired one-tailed t-test with performance across the $n$=3 random seeds as trials.} 
\label{fig:results}
\end{figure}

To validate \methodnameabbrev{}, we ran 8 experiments that reflect different challenges for morphological profiling, including integration across batches and sources as well as generalization to microscopy batch effects and perturbations unseen in training (Fig~\ref{fig:results}. We benchmarked \methodnameabbrev{} against baselines and ablated the \contrastivenameabbrev{} loss on the predicted images -- and therefore information about chemical structures -- to evaluate how integration of chemical compounds improves representation learning in morphological profiling. Our results demonstrate that our full \methodnameabbrev{} method using the \contrastivenameabbrev{} loss on the predicted images results in a sizable improvement over the CellProfiler, SIMCLR, and CLIP baselines. While performance improvements over using just the \contrastivenameabbrev{} loss alone were smaller, and thus not always statistically significant for individual experiments, pooling experiments indicated that the \contrastivenameabbrev{} loss on the predicted images significantly improves performance (F-ratio value of $9.249$, p-value $0.0228$ with a one-way repeated measures ANOVA). This observation suggests that modeling chemical compounds as treatments during training is effective at improving representations of microscopy images. We also observe that across all settings, all self-supervised learning methods improve over a pre-trained ResNet encoder on ImageNet, suggesting that some form of self-supervised fine-tuning on microscopy images is beneficial to performance regardless of method. In subsequent sections, we expand on the results of each experiment. We report 1-nearest neighbor matching results in the main manuscript, with top-3 and top-5 results are available in \nameref{S1_Table}, \nameref{S2_Table}, \nameref{S3_Table}, and \nameref{S4_Table}.

\subsection*{\methodnameabbrev{} improves representations of images}

We first trained \methodnameabbrev{} and the baseline models on the ID (in-distribution) training dataset of the POS-CTL wells, on a total of 21,284 wells from 6 sources (Fig~\ref{fig:pos_control_stratification}B). The positive control wells represent 8 compounds nominated to be morphologically distinct from DMSO and from each other \cite{chandrasekaran2023jump}. We used this positive control dataset to measure if representations can identify reproducible effects of chemical compounds that induce relatively strong phenotypic effects across increasing batch effects. Using the trained models, as well as CellProfiler features, we extracted representations for 9,078 unseen wells (Fig \ref{fig:pos_control_stratification}B; a subset of these representations are shown as a UMAP in \nameref{S2_Fig}. We then measured compound-replicate matching accuracy, specifically how frequently the nearest-retrieved NSB (not in same batch) and NSS (not in same source) neighbor was treated with the same chemical compound (Fig~\ref{fig:results}A; \nameref{S5_Table}). CellProfiler features, the common-practice standard in the field, yielded a 1-NN NSB accuracy of 36.56\% and a 1-NN NSS accuracy of 9.32\% on the positive control test dataset, where expected accuracy of random guessing across 8 compounds is 12.5\%. In contrast, \methodnameabbrev{} achieved a 1-NN NSB accuracy of 62.24\% and a NSS accuracy of 28.43\%, significantly outperforming both CellProfiler features as well SimCLR, (37.77\% 1-NN NSB and 16.56\% NSS) and CLIP, (34.33\% 1-NN NSB and 12.16\% NSS) (Fig~\ref{fig:results}A; \nameref{S5_Table}). DINOv3, as a more state-of-the-art self-supervised learning method, performs better than SimCLR and CLIP (42.41\% 1-NN NSB and 19.21\% NSS), but is still outperformed by MICON. Finally, we observe that OpenPhenom-S/16, a microscopy foundation model, generally performs comparable to or worse than CellProfiler (35.41\% 1-NN NSB and 8.22\% NSS).

We then asked whether these performance gains were due to \methodnameabbrev{}'s multimodal integration of chemical structures or if they could be fully attributed to the \contrastivenameabbrev{} component, which operates exclusively on images. To test this, we ablated the part of the \methodnameabbrev{} loss on the predicted images, which is the only component that integrates chemical compounds, and trained a model using just the \contrastivenameabbrev{} loss. While this ablated model underperformed the full \methodnameabbrev{} model, it still outperformed the baselines, achieving a 1-NN NSB accuracy of 56.57\% and a 1-NN NSS accuracy of 25.21\% (Fig~\ref{fig:results}A; \nameref{S5_Table}). Together, these results suggest that, while the majority of \methodnameabbrev{}'s strong performance is driven by the \contrastivenameabbrev{} loss, the integration of chemical compound information further improved the quality of its representations. 

\subsection*{\methodnameabbrev{} outperforms baseline methods on unseen sources}

We next sought to understand if \methodnameabbrev{} generalized to sources unseen in training. Many representation learning methods are designed to learn representations invariant to microscopy batch effects. This is true not just of \methodnameabbrev{}, whose contrastive method minimizes distances between images treated with the same perturbation across different sources, but also CLIP-based approaches \cite{sanchez2023cloome, fradkin2024molecules}, which map images treated with the same perturbation across sources to the same chemical compound representation. Hence, a major question with these representation learning methods is if these learned invariances generalize, or if they are overfit to sources seen during training. 

To assess generalization to unseen sources, we designed an OOD (out-of-distribution) experiment using the POS-CTL wells, in which the training dataset consists of $5$ seen sources and the \nntest{} dataset consists wholly of a sixth source unseen during training (Fig~\ref{fig:pos_control_stratification}C). For each representation learning method, we evaluated compound-replicate matching performance on the unseen source test dataset. We found that all representation learning methods dropped in performance when sources are unseen in training (Fig~\ref{fig:results}C; \nameref{S5_Table}). In the NSB setting, SimCLR (34.45\%), CLIP (32.53\%), and DINOv3 (33.54\%) no longer outperformed CellProfiler (39.26\%), suggesting that performance above CellProfiler may be due to overfitting to seen sources. However, \methodnameabbrev{} (51.57\%) still significantly outperformed CellProfiler, SimCLR, and CLIP, indicating \methodnameabbrev{}'s capacity to generalize to unseen sources. In the NSS setting, \methodnameabbrev{}'s performance (11.25\%) was not significantly better than that of SimCLR (9.14\%), CLIP (10.76\%), or DINOv3 (10.97\%), although \methodnameabbrev{} was significantly better than CellProfiler (7.32\%). We note that the unseen source NSS experiment is the only setting where all methods underperformed random guessing, suggesting that generalization to a source unseen during training remains an unsolved problem.

\subsection*{\methodnameabbrev{} demonstrates robust performance over larger compound datasets with unseen compounds}

While our previous experiments focused on data representing just 8 chemical compound treatments, most morphological profiling experiments assay a greater number of perturbations. Therefore, we next sought to evaluate the performance of \methodnameabbrev{} when trained on datasets with a larger number of chemical compounds. We trained a \methodnameabbrev{} model on a total of 35,156 wells subsetted from the Target-2 plates, which image 301 perturbations replicated across all sources (Fig~\ref{fig:tgt2_stratification}B; a subset of these representations are shown as a UMAP in \nameref{S3_Fig}).

As with our POS-CTL well experiments, we evaluated NSB and NSS compound-replicate matching accuracy on an unseen \nntest{} dataset. Due to the greater number of chemical compounds, this task becomes much more challenging -- the accuracy expected from random guessing is 0.33\% (1 in 301). The CellProfiler, SimCLR, CLIP, DINOv3, and OpenPhenom-S/16 baselines exhibited performance ranging from $4.26$ -- $8.45\%$ for NSB accuracy and from $0.21$ -- $3.34\%$ for NSS accuracy, while \methodnameabbrev{} significantly improved on these baselines, providing $11.89\%$ NSB accuracy and $3.93\%$ NSS accuracy (Fig~\ref{fig:results}B). We show examples of the images retrieved by MICON versus CellProfiler in \nameref{S4_Fig}.

One limitation of our compound-replicate evaluation is that representation learning strategies that incorporate information about chemical compounds (i.e., both \methodnameabbrev{} and CLIP) may potentially overfit to compounds seen during training, which may result in a poorer representation of phenotypes induced by new, unseen compounds.
To test this, we constructed an independent, OOD test dataset of wells treated with $184$ compounds unseen during training (Fig~\ref{fig:tgt2_stratification}C). We sampled these wells from the Compound plates of the JUMP-CP cpg0016 dataset and evaluated whether models trained on the Target-2 plates could still identify reproducible effects in images across these unseen compounds. To make this retrieval task more challenging, we allowed our nearest neighbor classifiers to retrieve for unseen compounds, compounds seen during training. We did this by curating a \nntest{} set that only consists of the $184$ seen compounds, but including both seen and unseen compounds in the \nntrain{} dataset (for a total of $485$ compounds, $301$ seen and $184$ unseen).

We observed that the baseline representation learning methods largely were not robust on compounds unseen in the training dataset, resulting in accuracies mostly well below that of CellProfiler features, despite being originally superior on seen compounds (Fig~\ref{fig:results}D). While CellProfiler features had an NSB accuracy of 3.53\% and an NSS of 1.66\%, SimCLR resulted in an NSB of 2.17\% and an NSS of 0.72\%, CLIP had an NSB of 2.33\% and an NSS of 0.99\%, OpenPhenom had a NSB of 2.69\% and a NSS of 0.97\%, and DINOv3 had an NSB of 2.75\% and a NSS of 1.79\%. In contrast, \methodnameabbrev{} performed significantly better than all baselines on unseen compounds, achieving an NSB of 4.92\% and an NSS of 2.15\% (Fig~\ref{fig:results}D; \nameref{S6_Table}). These results demonstrate that incorporating information about perturbations into representation learning yields improvements generalization to new, unseen perturbations. We note unlike the out-of-distribution experiments with POS-CTL wells (which held out sources from training), we perform above random guessing with unseen compounds, suggesting that generalizing to unseen compounds is less challenging than generalizing to unseen sources.

\begin{figure}[!h]
\centering
\includegraphics[width=\textwidth]{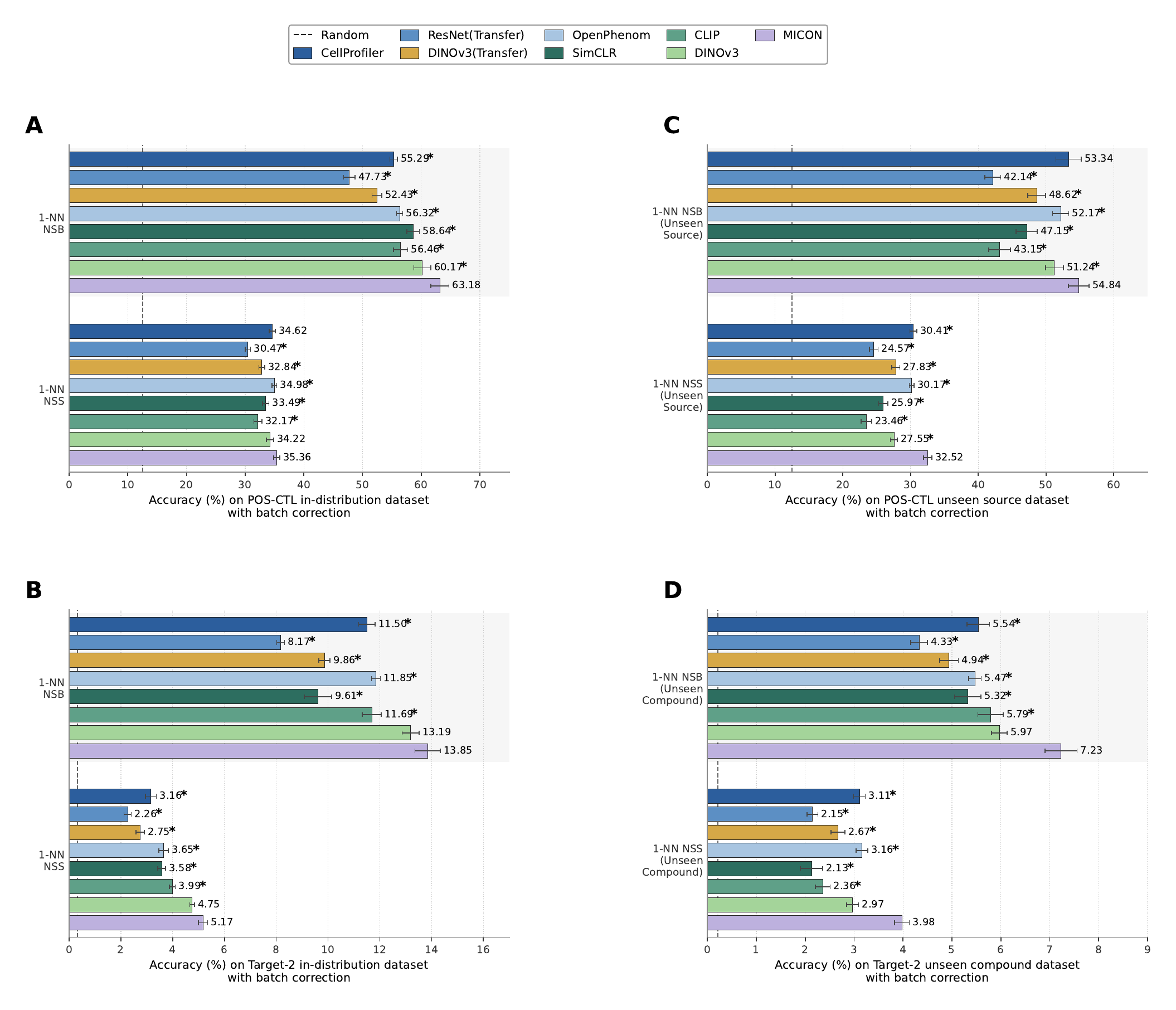}
\caption{\textbf{Compound-replicate matching accuracy for representation learning methods across evaluation settings, when microscopy batch effect correction is applied via spherizing.} Retrieval accuracies on the \textbf{(A)} POS-CTL ID \nntest{} dataset, \textbf{(B)} Target-2 ID \nntest{} dataset, \textbf{(C)} POS-CTL OOD \nntest{} dataset (unseen source), and \textbf{(D)} Target-2 OOD \nntest{} dataset (unseen compounds). NSB designates retrieval of the nearest neighbor not in the same batch; NSS designates retrieval of the nearest neighbor not in the same source. $n$=3 models trained and evaluated with different dataset stratifications and different random seeds; the mean accuracy is labeled, and the error bars represent standard deviation.. Asterisk indicates that \methodnameabbrev{} significantly outperforms the baseline, defined as $p<0.05$ on an unpaired one-tailed t-test with performance across the $n$=3 random seeds as trials.} 
\label{fig:batch_correction_results}
\end{figure}

\subsection*{\methodnameabbrev{} benefits from microscopy batch effect correction}

While our evaluations thus far used the raw representations from \methodnameabbrev{} and baselines, in practice, batch effect correction is often applied \textit{post hoc} to representations from morphological profiling. One common transformation is to spherize representations for each plate such that the negative control representations have an identity covariate matrix \cite{ando2017improving, way2022morphology}. \methodnameabbrev{} is designed to correct for batch effects through its training set-up, which enforces similarity between images with the same perturbation across different batches. We thus next sought to confirm that performance improvements from MICON were not because of trivial correction of batch effects, such that common adopted standards for batch effect correction could equalize the performance of baselines that do not inherently build in batch effect correction.


We assessed the impact of microscopy batch effect correction on \methodnameabbrev{} representations relative to the baseline methods by conducting compound-replicate matching, but with spherized representations, across the 8 evaluation settings (Fig~\ref{fig:batch_correction_results}; \nameref{S7_Table} and \nameref{S8_Table}). Spherizing substantially improved the performance of all representations evaluated. Interestingly, SimCLR, CLIP, OpenPhenom-S/16, and DINOv3 underperformed or performed similarly to CellProfiler in almost all experiments (except for the in-distribution Target-2 setting for DINOv3). These observations suggest that the previously observed improvements from the SimCLR, CLIP, OpenPhenom-S/16, and DINOv3 baselines without batch effect correction (Fig~\ref{fig:results}) may be driven by an inherent robustness to microscopy batch effects in the feature representations. Thus, this performance difference is eliminated when CellProfiler features are post-processed with batch effect correction (Fig~\ref{fig:batch_correction_results}). Critically, \methodnameabbrev{} representations still significantly outperformed all self-supervised representations in all experiments, and were significantly better or similar to CellProfiler features in all experiments (significantly better in 6 of 8 settings). These results suggest that \methodnameabbrev{} performance is driven by superior recognition of phenotypic effects, not just robustness to microscopy batch effects (Fig~\ref{fig:batch_correction_results}).

\subsection*{\methodnameabbrev{} achieves mechanism of action enrichment}

Although compound-replicate matching tests if representations are able to identify reproducible biological signal, it does not assess whether compounds are grouped sensibly in representation space (i.e., compounds that induce biologically similar phenotypes clustering together). We thus evaluated our models on mechanism of action (MoA) enrichment \cite{moshkov2024learning}, which tests if compounds with the same MoA are enriched in the top-ranked retrievals for a given query compound. 

\begin{table*}[!ht]
\centering
\begin{small}
\begin{sc}
\caption{\textbf{Mechanism of action enrichment for the cpg0016 compound dataset.} We report the average fold-of-enrichment with imputation (imputing the size of the background set as in \cite{moshkov2024learning}) and the percent of significant compounds.}
\resizebox{\textwidth}{!}{
\begin{tabular}{l|cccc}
\toprule
          & \makecell{Average Fold-of-Enrichment} & \makecell{Percent Significant Compounds}  \\
\midrule
        CellProfiler  & 20.69 & 11.45\% \\
        SimCLR  & 21.07 &  14.39\%\\
        CLIP  & 27.47 & 14.60\%\\
        DINOv3 & 35.39 & 15.97\%\\
        MICON & 37.52 & 16.18\%\\
\bottomrule
\end{tabular}}
\label{tab:cpg0016_moa_results}
\end{sc}
\end{small}
\end{table*}

We evaluated MoA enrichment on a curated subset of the cpg0016 compound plates, using spherized features extracted by each of our trained models plus CellProfiler features (Table \ref{tab:cpg0016_moa_results}). For the trained models we used the top performing replicate of each model from the Target-2 OOD setting. In addition to average fold-of-enrichment, we also report the percent of compounds with significant enrichment. The latter metric is more robust to when there is an infinite odds-ratio for some compounds, which occurs sometimes as there are few MoAs matched for some query compounds. While previous benchmarks imputed the size of the background set when this occurs \cite{moshkov2024learning}, this can lead to inflated differences in values when there are small differences in behavior of models.

We observe that MICON achieves a higher performance, both on average fold-of-enrichment and percent significant compounds than other methods, suggesting that the improvements in performance MICON achieved in compound-replicate retrieval also translate into better clustering of compounds with similar mechanisms of action. All self-supervised models (SimCLR, CLIP, DINOv3, and MICON) outperformed CellProfiler features by a significant margin by both metrics, unlike our compound-replicate matching experiments. This may be because unlike compound-replicate retrieval, MoA enrichment metrics allow for retrieval of compounds from the same batch, making it less challenging of a batch effect setting.

\begin{table*}[!ht]
\centering
\begin{small}
\begin{sc}
\caption{\textbf{Mechanism of action enrichment for the BBBC036 dataset.} We report the average fold-of-enrichment with imputation (imputing the size of the background set as in \cite{moshkov2024learning}) and the percent of significant compounds.}
\resizebox{\textwidth}{!}{
\begin{tabular}{l|cccc}
\toprule
          & \makecell{Average Fold-of-Enrichment} & \makecell{Percent Significant Compounds}  \\
\midrule
        CellProfiler  & 25.62 &  9.20\% \\
        SimCLR  & 32.83 &  11.23\%\\
        CLIP  & 31.14 & 11.59\%\\
        DINOv3 & 31.23 & 11.23\%\\
        MICON & 31.26 & 11.23\%\\
\bottomrule
\end{tabular}}
\label{tab:bbbc036_results}
\end{sc}
\end{small}
\end{table*}

Next, we sought to similarly investigate MoA enrichment on the out-of-distribution BBBC036 dataset. Conversely, Besides outperforming CellProfiler, all self-supervised methods performed comparably. Although SimCLR exhibited slightly higher performance on average fold-of-enrichment than the other methods, it had the same percentage of significant compounds to the other methods, suggesting this performance may be due to inflation from imputation. Supporting the observation that all self-supervised learning methods perform similarly, we examined the overlap in significant compounds between methods (\nameref{S5_Fig}). All self-supervised methods retrieved compounds with the same MoAs for almost the same compounds. CellProfiler retrieved same-MoA compounds for a set of query compounds distinct from any self-supervised method. Overall, this suggests that MICON's improvements in retrieving biological signal is more restricted to in-distribution datasets.

\subsection*{\methodnameabbrev{} can synthesize phenotypic effects of unseen compounds}

The \contrastivenameabbrev{} loss on the predicted images means that \methodnameabbrev{} models should additionally be able to predict the phenotypic effect of compound perturbations. As a proof-of-concept test of this capability, we assessed how frequently the predicted representations from \methodnameabbrev{} models could retrieve a representation from a real image of cells perturbed with the same compound. We replicated each of the previous retrieval experiments, except using predicted representations instead of real ones. For each real perturbation representation originally used in compound-replicate matching, we replaced that representation with a predicted representation, using the spatially-nearest negative control as the basis for generation. This set-up enabled direct comparison of retrieval performance with predicted representations versus real representations. 

\begin{table*}[!ht]
\centering
\begin{small}
\begin{sc}
\resizebox{0.9\textwidth}{!}{
\begin{tabular}{l|cccc}
\toprule
         POS-CTL & \makecell{ID NSB} & \makecell{ID NSS} & \makecell{Unseen Sources NSB} & \makecell{Unseen Sources NSS}  \\
\midrule
        Random &  \multicolumn{4}{c}{12.50} \\
        MICON (Real Images) & $63.18 \pm 3.53$ & $35.36 \pm 3.21$ & $54.84 \pm 1.94$ & $32.52 \pm 2.32$\\
        MICON (Generated) & $54.75 \pm 3.21$ & $26.14 \pm 3.37$ & $47.26 \pm 3.46$  & $25.93 \pm 2.02$\\
\bottomrule
\toprule
          TGT-2 & \makecell{ID NSB} & \makecell{ID NSS} & \makecell{Unseen Compounds NSB} & \makecell{Unseen Compounds NSS}  \\
\midrule
        Random &  \multicolumn{4}{c}{0.33} \\
        MICON (Real Images) & $13.85 \pm 2.79$ & $5.17 \pm 1.28$ & $7.23 \pm 1.73$ & $3.98 \pm 0.67$ \\
        MICON (Generated) & $13.64 \pm 2.03$ & $4.76 \pm 0.93$ & $2.54 \pm  0.58$  & $1.07 \pm 0.21$ \\
\bottomrule
\end{tabular}}
\caption{\textbf{Nearest neighbor retrieval accuracies (\%) using real perturbation representations vs.\ generated counterfactual representations, across evaluation settings.} 
NSB designates retrieval of the nearest neighbor not in the same batch; NSS designates retrieval of the nearest neighbor not in the same source. Error represents the standard deviation across 3 independent replicates.}
\label{table:generated}
\end{sc}
\end{small}
\end{table*}

We observed that, for seen compounds, retrieval with the predicted representations only slightly declined in performance compared to with the real perturbation representations (Table~\ref{table:generated}), even when retrieving across sources and with unseen sources. For compounds unseen in training, the decline in performance was more substantial, but, notably, performance with predicted representations was still several folds higher than the accuracy from random guessing. These results suggest that \methodnameabbrev{}'s predicted representations capture phenotypic effects of compounds seen in training, with some ability to capture the phenotypic impact of completely unseen compounds. However, we caveat that the overall performance is low especially for unseen compounds, suggesting that \methodnameabbrev{} is likely not useful for prediction yet in its current state.

\section*{Discussion}
In this work, we introduced \methodnameabbrev{}, a molecular-image representation learning method for morphological profiling. \methodnameabbrev{} integrates two losses during training using contrastive learning: a perturbation-aware contrastive (\contrastivenameabbrev{}) loss that encourages representations of real images to be invariant to microscopy batch effects, and a predicted \contrastivenameabbrev{} loss that aligns representations of synthetic images synthesized using chemical compound information with those of real images of cells treated with the compound. Both components substantially advance representation learning. Even stand-alone, \methodnameabbrev{}'s \contrastivenameabbrev{} loss enables identification of reproducible phenotypic effects of compounds under batch effects, achieving improvements over both CellProfiler features and state-of-the-art representation learning methods in computer vision. We attribute this to the use of perturbation and microscopy batch metadata to mine both positive and negative samples for representation learning.

We showed that \methodnameabbrev{}'s\contrastivenameabbrev{} loss on the predicted images further improves performance in identifying phenotypic effects in images relative to models trained with just the \contrastivenameabbrev{} loss. This result demonstrates that integrating multimodal information about perturbation treatments can improve uni-modal representations of images, establishing directions for future morphological profiling methods. We also demonstrated that \textit{how} images and chemical compounds should be integrated is crucial. \methodnameabbrev{} models chemical compounds as a transformation of images, and this approach strongly outperforms the CLIP-like strategy of directly aligning images and chemical compounds in a shared embedding space.

Our evaluation also sheds novel insights into representation learning for cell painting. Taking advantage of the large-scale and multi-source nature of the JUMP-CP dataset, we designed evaluation splits that allow us to assess if morphological profiling methods can maintain performance in realistic application scenarios. Specifically, we evaluated performance when comparing images from different data-generating sources, and when sources or compounds are unseen during representation learning (Fig~\ref{fig:pos_control_stratification},~\ref{fig:tgt2_stratification}). While evaluations that stratify across sources have been designed to benchmark batch effect correction methods \cite{arevalo2024evaluating}, our study systematically benchmarked different representation learning methods in a controlled way, using the same architectures and training data. Previous evaluations of representation learning typically only use datasets collected by a single source and do not stratify compounds/sources such that some are unseen during training \cite{tang2024morphological, sanchez2023cloome, doron2023unbiased}. Here, our benchmarks exposed that some representation learning methods previously reported to be effective for morphological profiling may be overfitting and thus overestimating their performance. For example, while SimCLR outperforms CellProfiler features when representing seen compounds, it underperforms CellProfiler on compounds unseen during training. In contrast, \methodnameabbrev{} is robust and demonstrates consistent improvements over both CellProfiler and baseline representation learning methods across all scenarios evaluated. \methodnameabbrev{}'s improvements hold even when compounds are unseen in training and even when considering more challenging batch effects like differing imaging sources.

Future work should explore if \methodnameabbrev{} stays robust when deployed on larger and less curated datasets. In this work, we focused on a small number of perturbations that were more likely to have strong phenotypic effects. However, drug screening experiments are usually systematic in nature and will thus include many compounds that elicit minimal phenotypic impact on cells relative to negative controls. This impacts both training and evaluation. In training, as \methodnameabbrev{}'s \contrastivenameabbrev{} loss forces the model to distinguish compound-treated images from negative controls, it is not clear if scaling the number of compounds in the training dataset will adversely affect learning. Future work should analyze the trade-off between indiscriminately scaling the number of images and/or compounds versus downsampling datasets to just the stronger phenotypic effects. For evaluation, curating compounds that have minimal but reproducible effects without risking selecting compounds that have no reproducible phenotypic impacts on cells is challenging, but we think evaluating the sensitivity of models to more subtle phenotypes is crucial in future work. In this work, we focused on the POS-CTL wells and the Target-2 plates of cpg0016, which are reproduced multiple times by multiple sources. In contrast, some compounds in cpg0016 are only assayed by a subset of sources and once by each source. This reflects another trade-off in training data that should be explored in future work. Namely, it is unclear if focusing training on a small subset of well-reproduced compounds can explain some of the effectiveness of \methodnameabbrev{} or if expanding the number of compounds at the cost of less coverage across sources would further improve our method. Finally, there are opportunities in interpretation that we have not explored in this work: future work should investigate if the compound encoder attends to parts of chemical compounds that are responsible for biological mechanism of action.

While we demonstrated that \methodnameabbrev{} is capable of synthesizing the phenotypic effects of some unseen compounds as a consequence of its pretraining task, our evaluation was an initial proof-of-concept to determine if any predictive signal was present. It is unclear the effectiveness of this strategy compared to other proposals like CLIP-based strategies \cite{sanchez2023cloome, fradkin2024molecules}, and future work should do a comparative analysis. A more thorough exploration of hyperparameters could further advance our model's capabilities. In our experiments, we equally weighted the \contrastivenameabbrev{} losses on predicted versus real images. We demonstrated that although we have some signal in synthesizing the phenotypic effects of unseen compounds, our performance is weak, and one possibility is that in study, we dedicated too little of our loss to generating predictions. Increasing the weight on this loss, either throughout training or during later epochs, may enhance performance in producing synthetic image representations. 

Our current proposed model is restricted to chemical compounds as input perturbations. Extending \methodnameabbrev{} to additional perturbation modalities, such as genetic or environmental perturbations, could similarly model them as treatments that induce a transformation of images. \methodnameabbrev{}'s framework enables prediction of representations and, hence, of the phenotypic effects of unseen compounds. While optimizing this capability is beyond the scope of our work, the ability to predict the phenotypic effects of perturbations prior to experimentation could make high-throughput drug screens more resource efficient by triaging compounds that are more likely to be effective. Towards this capability, our proof-of-concept demonstrates that generation is possible directly in a learned latent space, paralleling similar work with conditional generative models \cite{palma2023predicting} that produce synthetic images that reflect perturbation phenotypes. In sum, through multimodal modeling of perturbations as treatments on images, MICON provides a new direction for representation learning in morphological profiling, paving the way for new methods that explicitly leverage the multimodal nature of microscopy screens.

\FloatBarrier

\section*{Acknowledgments}
We thank Sean Whitzell for assistance in dataset storage and management.

\section*{Supporting information}

\paragraph*{S1 Text}
\label{S1_Text}
\textbf{Supplementary Methods.} Supplementary methods describing how permutation tests for mechanism of action enrichment were conducted.

\paragraph*{S1 Table}
\label{S1_Table}
\textbf{Top-1/3/5 retrieval accuracy for the POS-CTL test datasets.} We report not-same-batch (NSB) and not-same-source (NSB) across the 4 benchmarking settings. For top-3 and top-5 retrieval, we score the retrieval as correct if any samples retrieved in the top 3 or 5 retrievals are the same compound as the query. Values are means across n=3 replicates with random seeds.

\paragraph*{S2 Table}
\label{S2_Table}
\textbf{Top-1/3/5 retrieval accuracy for the TGT2 test datasets.} We report not-same-batch (NSB) and not-same-source (NSB) across the 4 benchmarking settings. For top-3 and top-5 retrieval, we score the retrieval as correct if any samples retrieved in the top 3 or 5 retrievals are the same compound as the query. Values are means across n=3 replicates with random seeds.

\paragraph*{S3 Table}
\label{S3_Table}
\textbf{Top-1/3/5 retrieval accuracy for the POS-CTL test datasets after post-processing with batch correction.} We report not-same-batch (NSB) and not-same-source (NSB) across the 4 benchmarking settings. For top-3 and top-5 retrieval, we score the retrieval as correct if any samples retrieved in the top 3 or 5 retrievals are the same compound as the query. Values are means across n=3 replicates with random seeds.

\paragraph*{S4 Table}
\label{S4_Table}
\textbf{Top-1/3/5 retrieval accuracy for the TGT2 test datasets after post-processing with batch correction.} We report not-same-batch (NSB) and not-same-source (NSB) across the 4 benchmarking settings. For top-3 and top-5 retrieval, we score the retrieval as correct if any samples retrieved in the top 3 or 5 retrievals are the same compound as the query. Values are means across n=3 replicates with random seeds.

\paragraph*{S5 Table}
\label{S5_Table}
\textbf{Top-1 Not-Same-Batch (NSB) / Not-Same-Source (NSS) 1-NN Retrieval Accuracy across the 4 benchmarking settings for the POS-CTL test datasets}, identical to the results in Figure \ref{fig:results} and reproduced in numerical form for reproducibility purposes. Error bars represent the standard deviation across $n=3$ replicates with random seeds. Asterisk indicates that MICON significantly outperforms the baseline, defined as $p < 0.05$ on an unpaired one-tailed t-test with performance across the $n=3$ random seeds as trials.

\paragraph*{S6 Table}
\label{S6_Table}
\textbf{Top-1 Not-Same-Batch (NSB) / Not-Same-Source (NSS) 1-NN Retrieval Accuracy across the 4 benchmarking settings for the TGT2 test datasets}, identical to the results in Figure \ref{fig:results} and reproduced in numerical form for reproducibility purposes. Error bars represent the standard deviation across $n=3$ replicates with random seeds. Asterisk indicates that MICON significantly outperforms the baseline, defined as $p < 0.05$ on an unpaired one-tailed t-test with performance across the $n=3$ random seeds as trials.

\paragraph*{S7 Table}
\label{S7_Table}
\textbf{Top-1 Not-Same-Batch (NSB) / Not-Same-Source (NSS) 1-NN Retrieval Accuracy across the 4 benchmarking settings for the POS-CTL datasets after batch correction post-processing}, identical to the results in Figure \ref{fig:results} and reproduced in numerical form for reproducibility purposes. Error bars represent the standard deviation across $n=3$ replicates with random seeds. Asterisk indicates that MICON significantly outperforms the baseline, defined as $p < 0.05$ on an unpaired one-tailed t-test with performance across the $n=3$ random seeds as trials.

\paragraph*{S8 Table}
\label{S8_Table}
\textbf{Top-1 Not-Same-Batch (NSB) / Not-Same-Source (NSS) 1-NN Retrieval Accuracy across the 4 benchmarking settings for the TGT-2 datasets after batch correction post-processing}, identical to the results in Figure \ref{fig:results} and reproduced in numerical form for reproducibility purposes. Error bars represent the standard deviation across $n=3$ replicates with random seeds. Asterisk indicates that MICON significantly outperforms the baseline, defined as $p < 0.05$ on an unpaired one-tailed t-test with performance across the $n=3$ random seeds as trials.

\paragraph*{S9 Table}
\label{S9_Table}
\textbf{Detailed metadata statistics for the POS-CTL data splits as shown in Figure \ref{fig:pos_control_stratification}}.

\paragraph*{S10 Table}
\label{S10_Table}
\textbf{Detailed metadata statistics for the TGT-2 data splits as shown in Figure \ref{fig:tgt2_stratification}}.

\paragraph*{S11 Table}
\label{S11_Table}
\textbf{Hyperparameters of the MICON architecture}.

\paragraph*{S1 Fig}
\label{S1_Fig}
\textbf{Heat map showing cosine similarity between aggregated well-level profiles of CellProfiler features across negative control wells for each batch.} Batches (x and y-axes) are labeled by their source in CP JUMP.

\paragraph*{S2 Fig}
\label{S2_Fig}
\textbf{Image representations from CellProfiler, SimCLR, CLIP, and MICON for the POS-CTL dataset, reduced using UMAP and visualized as a 2D scatterplot.} For each representation method, the top plot is colored by source, and the bottom plot is colored by perturbation, to visualize robustness to batch effects and clustering of biological phenotype, respectively. $n=2,000$ randomly sampled images are shown.

\paragraph*{S3 Fig}
\label{S3_Fig} 
\textbf{Image representations from CellProfiler, SimCLR, CLIP, and MICON for a subset of the Target-2 dataset, reduced using UMAP and visualized as a 2D scatterplot.} For each representation method, the top plot is colored by source, and the bottom plot is colored by perturbation, to visualize robustness to batch effects and clustering of biological phenotype, respectively. 10 compounds with accuracy higher than 0\% across all methods are randomly sampled to avoid cluttering the visualization with too many labels. $n=2,000$ randomly sampled images are shown.

\paragraph*{S4 Fig}
\label{S4_Fig} 
\textbf{NSB/NSS retrieval results using images sampled from the Target-2 dataset.} Representations of the query image (first column) are used to retrieve other images in the dataset. Negative control shows a negative control from the same batch. CellProfiler-NSB and CellProfiler-NSS designate the nearest neighbor of the query image in CellProfiler features in a different batch and source respectively. MICON-NSB and MICON-NSS are these retrievals but using MICON features, while MICON-Gen-NSB and MICON-Gen-NSS are retrievals using predicted representations. Blue frame images are treated with the same perturbation and orange frames indicate the treatments are from different perturbations/negative control. Channels are colored so that DNA is shown in blue, ER is shown in green, AGP (Actin, Golgi, Plasma Membrane) in red, mitochondria in magneta, and RNA in cyan. Channels are rescaled to the full intensity range of the image. A representative crop is taken for each image.

\paragraph*{S5 Fig}
\label{S5_Fig} 
\textbf{Venn diagram showing overlap in compounds significantly enriched in retrieving compounds with the same mechanism of action for the BBBC036 dataset across methods.}

\bibliography{reference}

\section*{Supporting information}
\setcounter{table}{0}
\captionsetup[table]{name=S\arabic{table} Table, labelsep=space}
\renewcommand{\thetable}{}

\setcounter{figure}{0}
\captionsetup[figure]{name=S\arabic{figure} Figure, labelsep=space}
\renewcommand{\thefigure}{}

\subsection*{S1 Text: Supplementary Methods}
\subsubsection*{Permutation Tests for Mechanism of Action Enrichment}
To quantify retrieval performance, we measure whether the top-ranked matches for a query are significantly enriched for the same mechanism of action compared to non-hits. Specifically, for a retrieval set defined by a cut-off $k$, let $a$ be the number of matching class samples retrieved, $b$ be the number of non-matching samples retrieved, $c$ be the number of matching samples not retrieved, and $d$ be the number of non-matching samples not retrieved. The Odds Ratio (OR) is conventionally defined as:

\begin{equation}
    OR = \frac{a \cdot d}{b \cdot c}
\end{equation}

Besides reporting the raw OR, we also report the percentage of classes that achieve statistically significant enrichment. Significance is determined by comparing the observed Odds Ratio ($OR_{obs}$) against the Odds Ratios of randomly selected samples ($OR_{rand}$) of the same set size. Using $M=100$ random permutations, the $p$-value for the $j$-th query is calculated as:

\begin{equation}
    p_j = \frac{1}{M} \sum_{i=1}^{M} \mathcal{I}(OR_{rand}^{(i)} \geq OR_{obs})
\end{equation}

where $\mathcal{I}$ is the indicator function. Finally, we calculate the percentage of significant queries ($P_{sig}$) across the entire dataset. Let $N$ be the total number of queries. The metric is defined as the fraction of queries where the permutation test yields a $p$-value below the significance threshold $\alpha = 0.05$:

\begin{equation}
    P_{sig} = \frac{\sum_{j=1}^{N} \mathcal{I}(p_j < 0.05)}{N} \times 100
\end{equation}
\newpage


\begin{table*}[!ht]
\centering
\begin{small}
\begin{sc}
\caption{\textbf{Top-1/3/5} Not-Same-Batch (NSB) / Not-Same-Source (NSS) 1-NN Retrieval Accuracy across the 4 benchmarking settings for the POS-CTL test datasets}
\resizebox{\textwidth}{!}{
\begin{tabular}{l|cccc}
\toprule
         POS-CTL(ID) Top-1/3/5 & \makecell{1-NN NSB} & \makecell{1-NN NSS}  & \makecell{1-NN NSB (Unseen Source)} & \makecell{1-NN NSS (Unseen Source)}  \\
\midrule
        Random Chance &  \multicolumn{4}{c}{12.50} \\
        CellProfiler  & $36.6/38.2/40.7$  & $9.3/10.2/10.9$ & $39.3/41.2/43.5$  & $7.4/7.9/8.2$ \\
        ResNet(Transfer) & $34.7/35.2/37.5$  & $7.4/7.6/7.6$ & $33.0/33.5/36.2$  & $7.3/7.3/7.5$ \\
\midrule
        SimCLR  & $37.8/38.6/38.6$ & $16.6/16.9/16.9$  & $34.5/34.5/34.9$ & $9.1/9.1/9.3$  \\
        CLIP  & $34.3/34.7/35.5$ & $12.2/13.1/13.9$ & $32.5/33.6/33.9$ & $10.8/11.2/11.2$ \\
        DINOv3 & $42.4/43.6/44.9$ & $19.2/19.5/20.2$  & $33.5/33.9/34.3$ & $11.0/11.8/12.5$  \\
\midrule 
        MICON (BAC Loss Only) &  $56.6/58.2/58.9$ & $25.2/26.3/26.6$ & $48.1/49.2/49.7$ & $10.2/10.2/10.6$ \\
        MICON & $\mathbf{62.2/64.7/65.5}$ & $\mathbf{28.4/29.6/30.4}$ & $\mathbf{51.6/53.1/53.7}$ & $\mathbf{11.3/13.3/14.7}$ \\

\bottomrule
\end{tabular}}
\label{tab:topk_pos_ctl}
\end{sc}
\end{small}
\end{table*}

\begin{table*}[!ht]
\centering
\begin{small}
\begin{sc}
\caption{\textbf{Top-1/3/5 Not-Same-Batch (NSB) / Not-Same-Source (NSS) 1-NN retrieval accuracy across the 4 benchmarking settings for the TGT2 test datasets}}
\resizebox{\textwidth}{!}{
\begin{tabular}{l|cccc}
\toprule
         TGT-2 (ID) & \makecell{1-NN NSB (ID)} & \makecell{1-NN NSS (ID)}  & \makecell{1-NN NSB \\ (Unseen Compound)} & \makecell{1-NN NSS \\ (Unseen Compound)}  \\
\midrule
        Random Chance &  \multicolumn{2}{c}{0.33}& \multicolumn{2}{c}{0.21} \\
        CellProfiler  & $5.03/5.89/7.33 $ & $0.67/0.67/1.12 $  & $3.53/3.72/3.72 $ & $1.66/1.79/1.79 $  \\
        ResNet(Transfer) & $2.34/2.34/2.79 $  & $0.21/0.37/0.76 $ & $1.97/1.97/1.97 $  & $0.37/0.37/0.37 $ \\
\midrule
        SimCLR  & $5.61/5.61/6.38 $ & $1.42/1.42/1.42 $  & $2.17/2.17/2.35 $ & $0.42/0.42/0.42 $  \\
        CLIP  & $6.22/6.76/7.44 $ & $2.99/2.99/3.68 $  & $2.33/2.33/2.33 $ & $0.99/0.99/0.99 $ \\
        DINOv3  & $8.45/8.96/9.34 $ & $3.34/3.34/3.65 $  & $2.75/2.75/2.94 $ & $1.37/1.37/1.46 $ \\
\midrule 
        MICON (BAC Loss Only) &  $11.17/12.36/13.19 $ & $3.65/3.74/3.74 $ &  $ 3.76/3.76/3.76 $ & $1.79/1.79/2.35 $ \\
        MICON & $\mathbf{11.89/12.97/14.37}$ & $\mathbf{3.93/4.35/4.35}$ & $\mathbf{4.92/5.82/5.82}$ & $\mathbf{2.15/2.15/2.35}$ \\

\bottomrule
\end{tabular}}

\label{tab:topk_tgt2_results}
\end{sc}
\end{small}
\end{table*}

\begin{table*}[!ht]
\centering
\begin{small}
\begin{sc}
\caption{\textbf{Top-1/3/5 Not-Same-Batch (NSB) / Not-Same-Source (NSS)} 1-NN Retrieval Accuracy across the 4 benchmarking settings the  POS-CTL test datasets after batch correction post-processing}
\resizebox{\textwidth}{!}{
\begin{tabular}{l|cccc}
\toprule
         POS-CTL(ID) & \makecell{1-NN NSB} & \makecell{1-NN NSS}  & \makecell{1-NN NSB (Unseen Source)} & \makecell{1-NN NSS (Unseen Source)}  \\
\midrule
        Random Chance &  \multicolumn{4}{c}{12.50} \\
        CellProfiler  & $55.3/58.2/63.5$  & $34.6/37.5/40.1$ & $53.3/57.2/62.6$  & $30.4/33.9/36.7$ \\
        ResNet(Transfer) & $47.7/49.5/50.2$  & $30.4/32.6/34.1$ & $ 42.1/45.5/48.6$  & $24.6/27.3/31.6$ \\
\midrule
        SimCLR  & $58.64/60.6/64.2$ & $33.5/35.9/37.1$  & $47.2/50.4/55.4$ & $26.0/28.5/29.3$  \\
        CLIP  & $56.5/59.3/62.9$ & $32.2/35.1/36.9$ & $43.3/48.9/53.1$ & $23.5/25.2/28.4$ \\
        DINOv3 & $60.2/63.5/66.3$ & $34.2/37.8/39.2$ & $51.2/53.4/57.3$ & $27.6/30.0/35.5$ \\

\midrule 
        MICON & $\mathbf{63.2/65.1/68.5}$ & $\mathbf{35.4/38.5/39.4}$ & $\mathbf{54.9/55.9/58.2}$ & $\mathbf{32.5/33.2/35.9}$ \\

\bottomrule
\end{tabular}}

\label{tab:topk_pos_ctl_bn}
\end{sc}
\end{small}
\end{table*}

\begin{table*}[!ht]
\centering
\begin{small}
\begin{sc}
\caption{\textbf{Top-1 Not-Same-Batch (NSB) / Not-Same-Source (NSS) 1-NN retrieval accuracy across the 4 benchmarking settings for the TGT2 test datasets after batch correction post-processing}}
\resizebox{\textwidth}{!}{
\begin{tabular}{l|cccc}
\toprule
         TGT-2 (ID) & \makecell{1-NN NSB (ID)} & \makecell{1-NN NSS (ID)}  & \makecell{1-NN NSB \\ (Unseen Compound)} & \makecell{1-NN NSS \\ (Unseen Compound)}  \\
\midrule
        Random Chance &  \multicolumn{2}{c}{0.33}& \multicolumn{2}{c}{0.21} \\
        CellProfiler  & $11.5/12.36/14.75$ & $3.16/3.79/4.86$  & $5.54/6.37/8.51$ & $3.11/4.86/5.75$  \\
        ResNet(Transfer) & $8.17/10.32/10.86$  & $2.26/2.82/3.16$ & $ 4.33/4.76/5.54 $  & $2.15/2.15/3.73$ \\
\midrule
        SimCLR  & $9.61/10.56/12.73 $ & $3.58/3.97/4.64 $  & $5.32/5.54/5.79 $ & $2.13/3.11/3.73 $  \\
        CLIP  & $11.69/12.36/12.96 $ & $3.99/4.32/5.17 $  & $5.79/5.93/6.22 $ & $2.36/3.34/3.95 $ \\
        DINOv3 & $13.19/13.95/14.75 $ & $4.75/5.32/5.86 $ & $5.97/6.34/6.79 $ & $2.97/3.73/4.32 $\\
\midrule 
        MICON & $\mathbf{13.85/14.75/15.98 }$ & $\mathbf{5.17/5.86/6.65}$ & $\mathbf{7.23/7.52/8.16}$ & $\mathbf{3.98/4.86/5.96 }$ \\

\bottomrule
\end{tabular}}

\label{tab:topk_tgt2_results_bn}
\end{sc}
\end{small}
\end{table*}

\begin{table*}[!ht]
\centering
\begin{small}
\begin{sc}
\caption{\textbf{Top-1 Not-Same-Batch (NSB) / Not-Same-Source (NSS) 1-NN Retrieval Accuracy across the 4 benchmarking settings for the POS-CTL test datasets,} identical to the results in Figure 4 and reproduced in numerical form for reproducibility purposes. Error bars represent the standard deviation across $n$=3 replicates with random seeds. Asterisk indicates that MICON significantly outperforms the baseline, defined as $p<0.05$ on an unpaired one-tailed t-test with performance across the $n$=3 random seeds as trials.}
\resizebox{\textwidth}{!}{
\begin{tabular}{l|cccc}
\toprule
         POS-CTL(ID) & \makecell{1-NN NSB} & \makecell{1-NN NSS}  & \makecell{1-NN NSB (Unseen Source)} & \makecell{1-NN NSS (Unseen Source)}  \\
\midrule
        Random Chance &  \multicolumn{4}{c}{12.50} \\
        CellProfiler  & $36.56 \pm 1.16^{\ast}$  & $9.32 \pm 0.79^{\ast}$ & $39.26 \pm 1.17^{\ast}$  & $7.42 \pm 0.79^{\ast}$ \\
        ResNet(Transfer) & $34.73 \pm 1.33^{\ast}$  & $7.41 \pm 0.63^{\ast}$ & $32.95 \pm 1.07^{\ast}$  & $7.31 \pm 0.71^{\ast}$ \\
        DINOv3(Transfer) & $35.84 \pm 1.18^{\ast}$ & $8.47 \pm 0.62^{\ast}$ & $36.41 \pm 1.21^{\ast}$ & $7.36 \pm 0.74^{\ast}$ \\
        OpenPhenom-S/16 & $35.41 \pm 1.07^{\ast}$  & $8.22 \pm 0.51^{\ast}$ & $38.77 \pm 1.12^{\ast}$  & $7.42 \pm 0.63^{\ast}$ \\
\midrule
        SimCLR  & $37.77 \pm 1.18^{\ast}$ & $16.56 \pm 0.95^{\ast}$  & $34.45 \pm 1.32^{\ast}$ & $9.12 \pm 0.86$  \\
        CLIP  & $34.33 \pm 1.46^{\ast}$ & $12.16 \pm 0.88^{\ast}$ & $32.53 \pm 1.34^{\ast}$ & $10.76 \pm 1.04$ \\
        DINOv3 & $42.41 \pm 0.97^{\ast}$ & $19.21 \pm 0.78^{\ast}$  & $33.54 \pm 1.17^{\ast}$ & $10.97 \pm 0.71$  \\
\midrule 
        MICON (BAC Loss Only) &  $56.57 \pm 1.65^{\ast}$ & $25.21 \pm 1.16^{\ast}$ & $48.13 \pm 1.41^{\ast}$ & $10.23 \pm 1.26$ \\
        MICON & $\mathbf{62.24 \pm 1.41}$ & $\mathbf{28.43 \pm 1.05}$ & $\mathbf{51.57 \pm 1.66}$ & $\mathbf{11.25 \pm 1.18}$ \\

\bottomrule
\end{tabular}}

\label{tab:pos_ctl_results}
\end{sc}
\end{small}
\end{table*}

\begin{table*}[!ht]
\centering
\begin{small}
\begin{sc}
\caption{\textbf{Top-1 Not-Same-Batch (NSB) / Not-Same-Source (NSS) 1-NN retrieval accuracy across the 4 benchmarking settings for the TGT2 test datasets,} identical to the results in 4 and reproduced in numerical form for reproducibility purposes. Error bars represent the standard deviation across $n$=3 replicates with random seeds. Asterisk indicates that MICON significantly outperforms the baseline, defined as $p<0.05$ on an unpaired one-tailed t-test with performance across the $n$=3 random seeds as trials.}
\resizebox{\textwidth}{!}{
\begin{tabular}{l|cccc}
\toprule
         TGT-2 (ID) & \makecell{1-NN NSB (ID)} & \makecell{1-NN NSS (ID)}  & \makecell{1-NN NSB \\ (Unseen Compound)} & \makecell{1-NN NSS \\ (Unseen Compound)}  \\
\midrule
        Random Chance &  \multicolumn{2}{c}{0.33}& \multicolumn{2}{c}{0.21} \\
        CellProfiler  & $5.03 \pm 0.43^{\ast}$ & $0.67 \pm 0.17^{\ast}$  & $3.53 \pm 0.29^{\ast}$ & $1.66 \pm 0.29^{\ast}$  \\
        ResNet(Transfer) & $2.34 \pm 0.26^{\ast}$  & $0.21 \pm 0.03^{\ast}$ & $1.97 \pm 0.16^{\ast}$  & $0.37 \pm 0.02^{\ast}$ \\
        DINOv3(Transfer) & $3.82 \pm 0.31^{\ast}$ & $0.45 \pm 0.06^{\ast}$ & $2.61 \pm 0.21^{\ast}$ & $0.86 \pm 0.09^{\ast}$ \\    
        OpenPhenom-S/16 & $4.26 \pm 0.33^{\ast}$  & $0.56 \pm 0.04^{\ast}$ & $2.69 \pm 0.14^{\ast}$  & $0.96 \pm 0.07^{\ast}$ \\
\midrule
        SimCLR  & $5.61 \pm 0.75^{\ast}$ & $1.42 \pm 0.15^{\ast}$  & $2.17 \pm 0.27^{\ast}$ & $0.42 \pm 0.12^{\ast}$  \\
        CLIP  & $6.22 \pm 0.64^{\ast}$ & $2.99 \pm 0.36^{\ast}$  & $2.33 \pm 0.26^{\ast}$ & $0.99 \pm 0.16^{\ast}$ \\
        DINOv3  & $8.45 \pm 0.57^{\ast}$ & $3.34 \pm 0.22$  & $2.75 \pm 0.31^{\ast}$ & $1.37 \pm 0.19$ \\
\midrule 
        MICON (BAC Loss Only) &  $11.17 \pm 0.89 $ & $3.65 \pm 0.47$ &  $ 3.76 \pm 0.43$ & $1.79 \pm 0.25$ \\
        MICON & $\mathbf{11.89 \pm 1.04}$ & $\mathbf{3.93 \pm 0.55}$ & $\mathbf{4.92 \pm 0.65}$ & $\mathbf{2.15 \pm 0.33}$ \\

\bottomrule
\end{tabular}}

\label{tab:tgt2_results}
\end{sc}
\end{small}
\end{table*}

\begin{table*}[!ht]
\centering
\begin{small}
\begin{sc}
\caption{\textbf{Top-1 Not-Same-Batch (NSB) / Not-Same-Source (NSS) 1-NN Retrieval Accuracy across the 4 benchmarking settings for the POS-CTL test datasets after batch correction post-processing,} identical to the results in Figure 5and reproduced in numerical form for reproducibility purposes. Error bars represent the standard deviation across $n$=3 replicates with random seeds. Asterisk indicates that MICON significantly outperforms the baseline, defined as $p<0.05$ on an unpaired one-tailed t-test with performance across the $n$=3 random seeds as trials.}
\resizebox{\textwidth}{!}{
\begin{tabular}{l|cccc}
\toprule
         POS-CTL(ID) & \makecell{1-NN NSB} & \makecell{1-NN NSS}  & \makecell{1-NN NSB (Unseen Source)} & \makecell{1-NN NSS (Unseen Source)}  \\
\midrule
        Random Chance &  \multicolumn{4}{c}{12.50} \\
        CellProfiler  & $55.29 \pm 0.63^{\ast}$  & $34.62 \pm 0.52$ & $53.34 \pm 1.87$  & $30.41 \pm 0.52^{\ast}$ \\
        ResNet(Transfer) & $47.73 \pm 0.97^{\ast}$  & $30.47 \pm 0.44^{\ast}$ & $ 42.14\pm 1.17^{\ast}$  & $24.57 \pm 0.65^{\ast}$ \\
        DINOv3(Transfer) & $52.43 \pm 0.82^{\ast}$ & $32.84 \pm 0.49^{\ast}$ & $48.62 \pm 1.31^{\ast}$ & $27.83 \pm 0.58^{\ast}$ \\    
        OpenPhenom-S/16 & $56.32 \pm 0.52^{\ast}$  & $34.98\pm 0.41^{\ast}$ & $52.17 \pm 1.19^{\ast}$  & $30.17 \pm 0.34^{\ast}$ \\
\midrule
        SimCLR  & $58.64 \pm 1.06^{\ast}$ & $33.49 \pm 0.54^{\ast}$  & $47.15 \pm 1.55^{\ast}$ & $25.97 \pm 0.68^{\ast}$  \\
        CLIP  & $56.46 \pm 1.22^{\ast}$ & $32.17 \pm 0.68^{\ast}$ & $43.15 \pm 1.63^{\ast}$ & $23.46 \pm 0.78^{\ast}$ \\
        DINOv3 & $60.17 \pm 1.45^{\ast}$ & $34.22 \pm 0.62$ & $51.24 \pm 1.31^{\ast}$ & $27.55 \pm 0.52^{\ast}$ \\

\midrule 
        MICON & $\mathbf{63.18 \pm 1.53}$ & $\mathbf{35.36 \pm 0.51}$ & $\mathbf{54.84 \pm 1.54}$ & $\mathbf{32.52 \pm 0.61}$ \\

\bottomrule
\end{tabular}}

\label{tab:pos_ctl_results_bn}
\end{sc}
\end{small}
\end{table*}

\begin{table*}[!ht]
\centering
\begin{small}
\begin{sc}
\caption{\textbf{Top-1 Not-Same-Batch (NSB) / Not-Same-Source (NSS) 1-NN retrieval accuracy across the 4 benchmarking settings for the TGT2 test datasets after batch correction post-processing,} identical to the results shown in Figure 5 and reproduced in numerical form for reproducibility purposes. Error bars represent the standard deviation across $n$=3 replicates with random seeds. Asterisk indicates that MICON significantly outperforms the baseline, defined as $p<0.05$ on an unpaired one-tailed t-test with performance across the $n$=3 random seeds as trials.}
\resizebox{\textwidth}{!}{
\begin{tabular}{l|cccc}
\toprule
         TGT-2 (ID) & \makecell{1-NN NSB (ID)} & \makecell{1-NN NSS (ID)}  & \makecell{1-NN NSB \\ (Unseen Compound)} & \makecell{1-NN NSS \\ (Unseen Compound)}  \\
\midrule
        Random Chance &  \multicolumn{2}{c}{0.33}& \multicolumn{2}{c}{0.21} \\
        CellProfiler  & $11.5 \pm 0.32^{\ast}$ & $3.16 \pm 0.21^{\ast}$  & $5.54 \pm 0.23^{\ast}$ & $3.11 \pm 0.12^{\ast}$  \\
        ResNet(Transfer) & $8.17 \pm 0.15^{\ast}$  & $2.26 \pm 0.14^{\ast}$ & $ 4.33\pm 0.17^{\ast}$  & $2.15 \pm 0.11^{\ast}$ \\
        DINOv3(Transfer) & $9.86 \pm 0.21^{\ast}$ & $2.75 \pm 0.16^{\ast}$ & $4.94 \pm 0.19^{\ast}$ & $2.67 \pm 0.14^{\ast}$ \\    
        OpenPhenom-S/16 & $11.85 \pm 0.17^{\ast}$  & $3.65\pm 0.18^{\ast}$ & $5.47 \pm 0.13^{\ast}$  & $3.16\pm 0.12^{\ast}$ \\
\midrule
        SimCLR  & $9.61 \pm 0.53^{\ast}$ & $3.58 \pm 0.15^{\ast}$  & $5.32 \pm 0.27^{\ast}$ & $2.13 \pm 0.23^{\ast}$  \\
        CLIP  & $11.69 \pm 0.37^{\ast}$ & $3.99 \pm 0.11^{\ast}$  & $5.79 \pm 0.26^{\ast}$ & $2.36 \pm 0.15^{\ast}$ \\
        DINOv3 & $13.19 \pm 0.33$ & $4.75 \pm 0.09$ & $5.97 \pm 0.16$ & $2.97 \pm 0.12$\\
\midrule 
        MICON & $\mathbf{13.85 \pm 0.49}$ & $\mathbf{5.17 \pm 0.18}$ & $\mathbf{7.23 \pm 0.33}$ & $\mathbf{3.98 \pm 0.15}$ \\

\bottomrule
\end{tabular}}

\label{tab:tgt2_results_bn}
\end{sc}
\end{small}
\end{table*}

\begin{table*}[!h]
\begin{center}
\begin{sc}
\caption{Detailed metadata statistics for the POS-CTL data splits shown in Figure 1.\label{tab:pos_ctl_datasets}}

\resizebox{\textwidth}{!}{
\begin{tabular}{l|ccccccc}
\toprule
         & \makecell{Perturbations}  & \makecell{\# Sources} & \makecell{Source IDs}  & \makecell{Batches}  & \makecell{Plates} & \makecell{Wells} & \makecell{FOVs} \\
\midrule
ID Training Dataset
& 8  & 6 & 2,3,5,7,8,11  & 57  & 679 & 21284 & 395279\\
ID Validation Dataset
& 8 & 6 & 2,3,5,7,8,11 & 6 & 139 & 4305 & 89988  \\
ID Test Dataset
& 8 & 6 & 2,3,5,7,8,11 & 12 & 283 & 9078 & 192094  \\
\hline
\makecell[l]{Unseen Sources \\
OOD Training Dataset}
& 8 & 5 & 3,5,7,8,11 & 57 & 756 & 23752 & 479556\\
\makecell[l]{Unseen Sources \\
OOD Validation Dataset}
& 8 & 5 & 3,5,7,8,11 & 5 & 137 & 4254 & 92535\\
\makecell[l]{Unseen Sources \\
OOD Query Dataset}
& 8 & 1 & 2 & 8 & 133 & 4096 & 60853\\
\makecell[l]{Unseen Sources \\
OOD Retrieval Dataset}
& 8 & 1 & 2 & 5 & 86 & 2565 & 44417\\

\bottomrule
\end{tabular}}

\end{sc}
\end{center}
\end{table*}

\begin{table*}[!h]
\begin{center}
\begin{sc}
\caption{\textbf{Detailed metadata statistics for the Target-2 data splits as shown in Figure 2.}}
\label{tab:tgt2_datasets}

\resizebox{\textwidth}{!}{
\begin{tabular}{l|ccccccc}
\toprule
         & \makecell{Perturbations}  & \makecell{\# Sources} & \makecell{Source IDs}  & \makecell{Batches}  & \makecell{Plates} & \makecell{Wells} & \makecell{FOVs} \\
\midrule
ID Training Dataset
& 301  & 10 & 2,3,4,5,6,7,8,9,11,13   & 71  & 92 & 35156 & 325834 \\
ID Validation Dataset
& 301  & 10 & 2,3,4,5,6,7,8,9,11,13   & 10  & 15 & 5723 & 53341 \\
ID Test Dataset
& 301 & 10 & 2,3,4,5,6,7,8,9,11,13 & 20 & 28 & 10875 & 99781  \\
\hline
\makecell[l]{Unseen Compounds \\ 
OOD Training Dataset}
& 301 & 10 & 2,3,4,5,6,7,8,9,11,13 & 91 & 122 & 46667 & 432500\\
\makecell[l]{Unseen Compounds \\
OOD Validation Dataset}
& 301 & 10 & 2,3,4,5,6,7,8,9,11,13  & 10 & 13 & 5087 & 46456\\

\makecell[l]{Unseen Compounds \\
OOD Query Dataset}
& 184 & 8 & 2,3,5,6,7,8,9,11  & 70 & 324 & 493 & 4053\\
\makecell[l]{Unseen Compounds \\
OOD Retrieval Dataset}
& 184 & 8 & 2,3,5,6,7,8,9,11 & 54 & 222  & 368 & 3079\\
\bottomrule
\end{tabular}}

\end{sc}
\end{center}
\end{table*}

\begin{table}[!h]
\centering
\caption{\textbf{Hyperparameters of the MICON architecture.}}
\label{params}
\begin{tabular}{ll|ll}
\toprule
\multicolumn{2}{c}{Model} & \multicolumn{2}{c}{Training} \\ \hline
ECFP4 fingerprint dimension & 2048 &  optimizer & Adam\\  \hline
image encoder embedding size  & 1000 &  batch size & 64\\ \hline
image projector hidden size & 512 & learning rate & 1e-3   \\ \hline
molecule hidden size & 512 & warmup steps  & 2000  \\ \hline
fusion module hidden size & 512 & weight decay  & 1e-2 \\ \hline
final projected embedding size & 256 & gradient clipping & 1.0 \\ \hline
\bottomrule

\end{tabular}
\end{table}

\begin{figure}[!htb]
\centering
\centerline{\includegraphics[width=\textwidth]{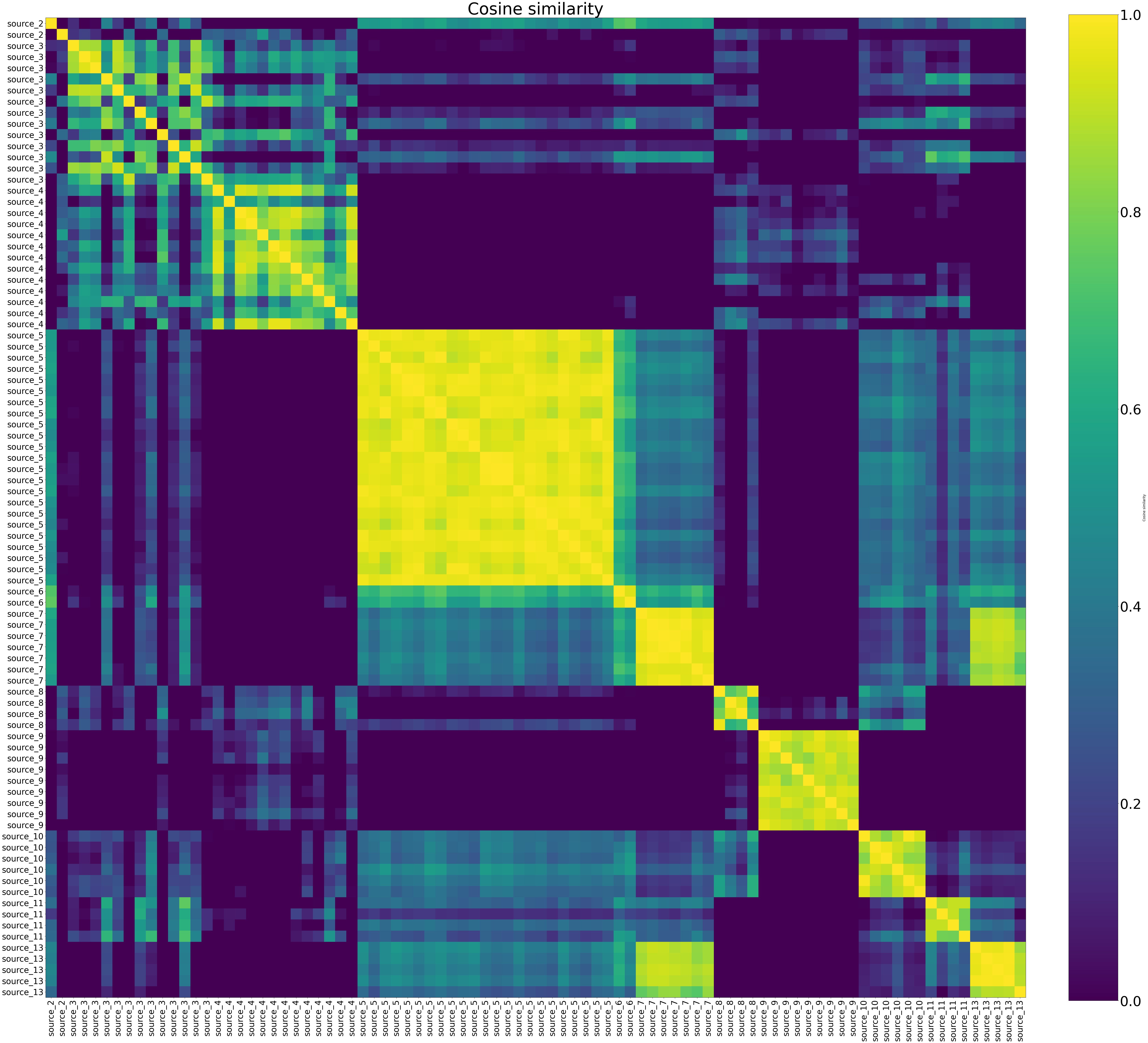}}
\caption{\textbf{Heat map showing cosine similarity between averaged CellProfiler features across negative control wells for each batch.} Batches (x and y-axes) are labeled by their source in CP JUMP.}
\label{fig:batch_effects}
\end{figure}

\begin{figure}[!htb]
\centering
\centerline{\includegraphics[width=\textwidth]{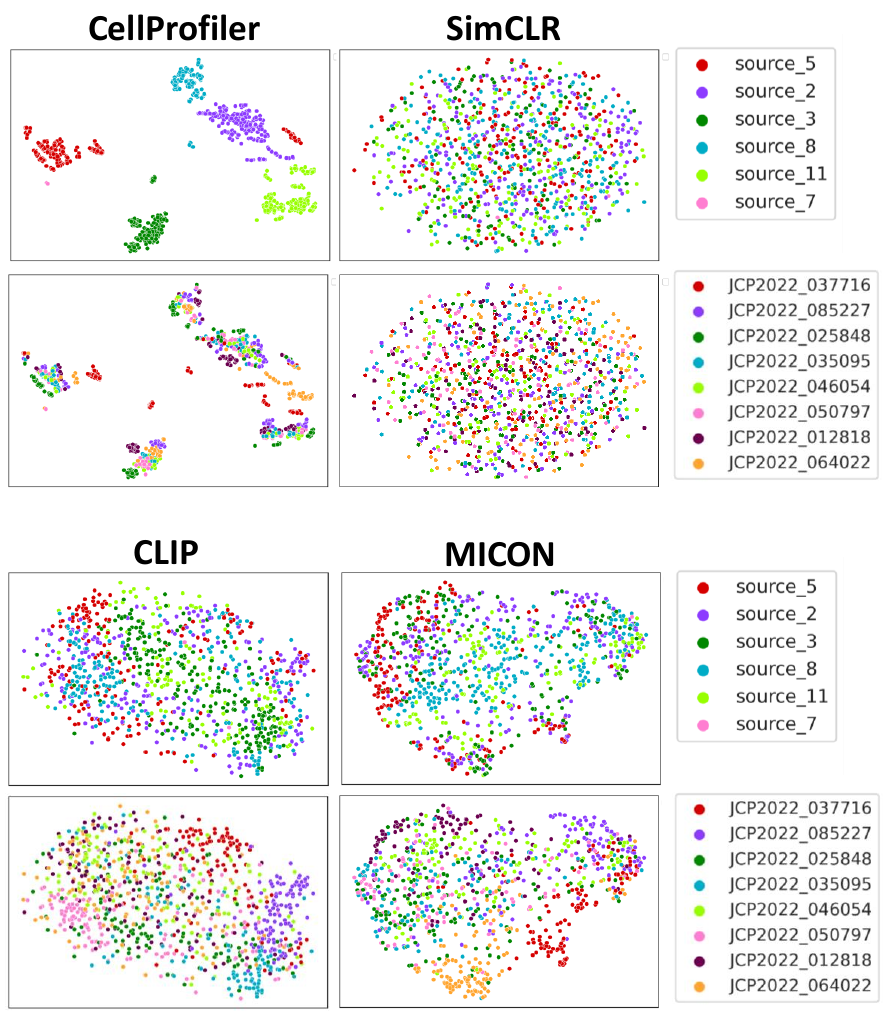}}
\caption{\textbf{Image representations from CellProfiler, SimCLR, CLIP, and \methodnameabbrev{} for the POS-CTL dataset, reduced using UMAP and visualized as a 2D scatterplot.} For each representation method, the top plot is colored by source, and the bottom plot is colored by perturbation, to visualize robustness to batch effects and clustering of biological phenotype, respectively. $n$=2,000 randomly sampled images are shown.}
\label{fig:umap_pos}
\end{figure}

\begin{figure}[!htb]
\centering
\centerline{\includegraphics[width=\textwidth]{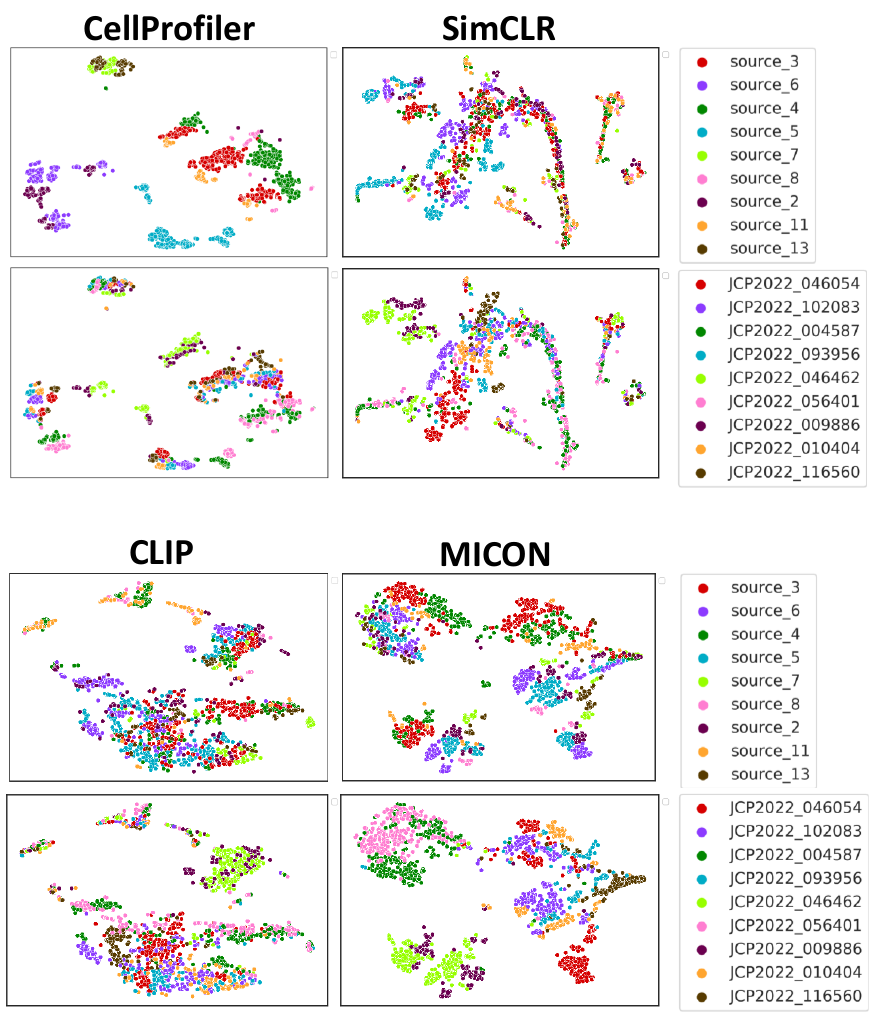}}
\caption{\textbf{Image representations from CellProfiler, SimCLR, CLIP, and \methodnameabbrev{} for a subset of the Target-2 dataset, reduced using UMAP and visualized as a 2D scatterplot.} For each representation method, the top plot is colored by source, and the bottom plot is colored by perturbation, to visualize robustness to batch effects and clustering of biological phenotype, respectively. 10 compounds with accuracy higher than 0\% across all methods are randomly sampled to avoid cluttering the visualization with too many labels. $n$=2,000 randomly sampled images are shown.}
\label{fig:umap_tgt2}
\end{figure}

\begin{figure}[!htb]
\centering
\centerline{\includegraphics[width=\textwidth]{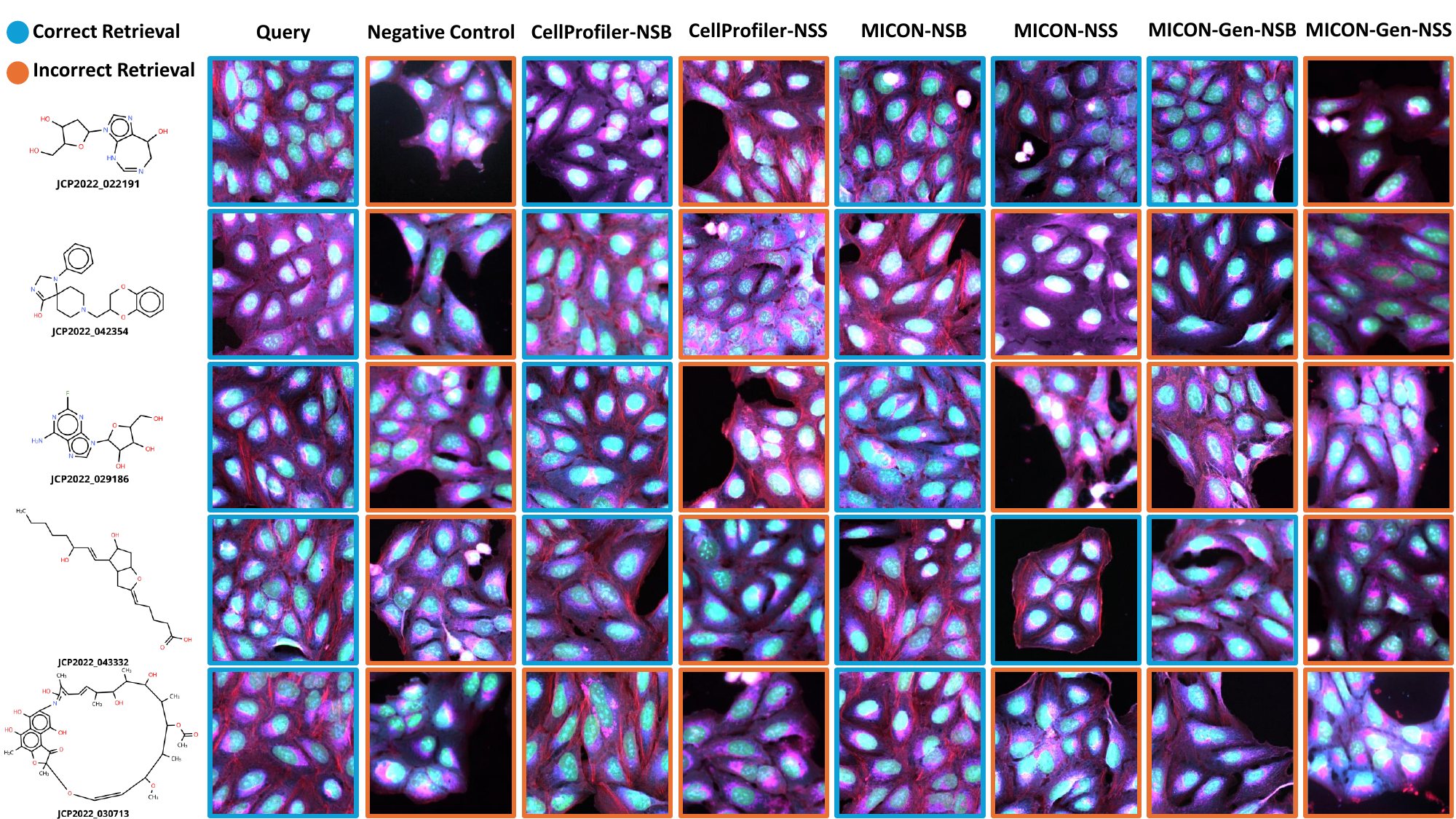}}
\caption{\textbf{NSB/NSS retrieval results using images sampled from the Target-2 dataset.} Representations of the query image (first column) are used to retrieve other images in the dataset. Negative control shows a negative control from the same batch. CellProfiler-NSB and CellProfiler-NSS designate the nearest neighbor of the query image in CellProfiler features in a different batch and source respectively. MICON-NSB and MICON-NSS are these retrievals but using MICON features, while MICON-Gen-NSB and MICON-Gen-NSS are retrievals using predicted representations. Blue frame images are treated with the same perturbation and orange frames indicate the treatments are from different perturbations/negative control. Channels are colored so that DNA is shown in blue, ER is shown in green, AGP (Actin, Golgi, Plasma Membrane) in red, mitochondria in magneta, and RNA in cyan. Channels are rescaled to the full intensity range of the image. A representative crop is taken for each image.}
\label{fig:retrieved_images}
\end{figure}

\begin{figure}[!htb]
\centering
\centerline{\includegraphics[width=\textwidth]{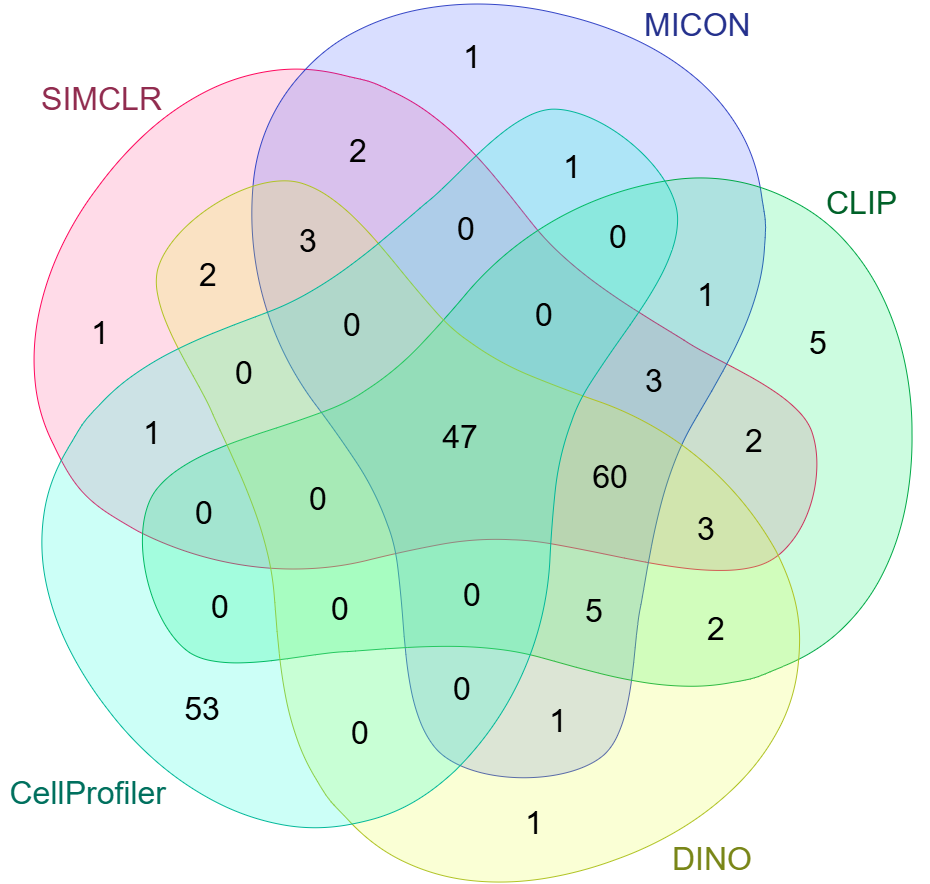}}
\caption{\textbf{Venn diagram showing overlap in compounds significantly enriched in retrieving compounds with the same mechanism of action for the BBBC036 dataset across methods.}}
\label{fig:sig_compounds}
\end{figure}

\end{document}